\PassOptionsToPackage{numbers}{natbib}
\documentclass{article}

\usepackage[preprint]{neurips_2025}

\usepackage[utf8]{inputenc} 
\usepackage[T1]{fontenc}    
\usepackage{hyperref}       
\usepackage{url}            
\usepackage{booktabs}       
\usepackage{amsfonts}       
\usepackage{nicefrac}       
\usepackage{microtype}      
\usepackage{xcolor}         
\usepackage{amsmath} 
\usepackage{hyperref}
\usepackage{graphicx}
\usepackage{algorithm}
\usepackage{algpseudocode}
\usepackage{enumerate}
\usepackage{titletoc}
\usepackage[page,header]{appendix}

\title{Lyapunov-Stable Adaptive Control for Multimodal Concept Drift}

\author{%
  Tianyu Bell Pan, Mengdi Zhu, Alexa Cole, Ronald Wilson, Damon Woodard
    \\
  Department of Electrical and Computer Engineering\\
  Florida Institute of National Security \\
  Applied Artificial Intelligence Group \\
  University of Florida\\
  Gainesville, FL 32611 \\
  \texttt{\{tpan1, zhum, alexa.cole, ronaldwilson\}@ufl.edu} \\
  \texttt{dwoodard@ece.ufl.edu} \\
}

\begin{document}

\maketitle

\begin{abstract}
  Multimodal learning systems often struggle in non-stationary environments due to concept drift, where changing data distributions can degrade performance. Modality-specific drifts and the lack of mechanisms for continuous, stable adaptation compound this challenge. This paper introduces LS-OGD, a novel adaptive control framework for robust multimodal learning in the presence of concept drift. LS-OGD uses an online controller that dynamically adjusts the model's learning rate and the fusion weights between different data modalities in response to detected drift and evolving prediction errors. We prove that under bounded drift conditions, the LS-OGD system's prediction error is uniformly ultimately bounded and converges to zero if the drift ceases. Additionally, we demonstrate that the adaptive fusion strategy effectively isolates and mitigates the impact of severe modality-specific drift, thereby ensuring system resilience and fault tolerance. These theoretical guarantees establish a principled foundation for developing reliable and continuously adapting multimodal learning systems. 
\end{abstract}

\section{Introduction}
Multimodal learning combines information from various data modalities, such as text and images, to enhance understanding and predictions. Real-world applications are naturally multimodal, such as autonomous systems leveraging cameras, LiDAR, and IMUs~\cite{ma2022compass}; driver monitoring systems integrating visual and physiological data~\cite{dontoh2025visual}; human-machine interaction using EOG and speech~\cite{li2024multimodal}; and online platforms handling diverse media~\cite{liu2025exploring}. A significant challenge arises when these systems operate in non-stationary environments, where the underlying data distribution changes over time, known as concept drift. By definition, concept drift is a change in the statistical properties of the target data distribution over time~\cite{liu2017regional}. Practically, a model's past experiences may no longer be valid as new patterns, topics, or styles emerge.

Concept drift is prevalent in real-world streaming data. For example, seasonal changes require weather prediction models to be adjusted periodically; shifting user preferences impact recommender systems; and social media or news topics can "drift" over weeks or months~\cite{liu2017regional}. In multimodal misinformation detection, concept drift might appear as new vocabulary or slang in text, or new image memes and deepfakes in visuals that were not present in the training data. These shifts in distribution can degrade a model's performance if it fails to adapt. Traditional machine learning models typically assume a stationary data distribution; therefore, the model's error rates increase when drift occurs, and previously learned decision boundaries may become misaligned.

Despite the prevalence of concept drift, current multimodal learning systems largely lack mechanisms to adapt to data distribution changes continuously. Most research on adapting to concept drift has focused on unimodal data streams (such as purely numeric or textual data)~\cite{yang2024adapting}. Multimodal scenarios introduce new challenges that have been underexplored: (1) \textbf{Modality-specific drifts:} One modality may experience drift while others remain stable, for example, the content of images may change while the relevance of the text stays constant. This situation requires adaptive fusion strategies that are unnecessary in single-modality settings. Traditional fusion techniques (like early or late fusion) rely on fixed combination rules, which become suboptimal under drift~\cite{baltruvsaitis2018multimodal}. (2) \textbf{Detection complexity:} Changes can be subtle and vary across modalities, complicating drift detection. Standard drift detectors often monitor a single error or a distribution metric~\cite{liu2017regional}. In the context of multimodal data, the question arises: Which signal indicates drift? How can we determine whether the drift originates from the textual or visual channel? (3) \textbf{Lack of theoretical guarantees:} Adaptation mechanisms in machine learning are often heuristic. There is a need for Lyapunov stability guarantees, which are essential in control theory for maintaining stable adaptive systems; however, these principles have not been fully applied to multimodal machine learning dealing with concept drift. Traditional adaptive control theory provides tools to ensure an adaptive system remains stable~\cite{dai2021lyapunov}, but these tools need further exploration in multimodal machine learning. (4) \textbf{Computational constraints:} In real-world applications like content filtering systems, models must adapt in real-time without extensive retraining. Many existing concept drift solutions retrain a new model on recent data or reset parts of the model, which can be inefficient or impractical in live systems. There is a pressing need for mechanisms that can continually, safely, and efficiently adjust models as new data becomes available. 

This paper addresses the above challenges by proposing a novel framework for Lyapunov-stable adaptive multimodal learning under concept drift (LS-OGD). Our key contributions are:

\textbf{Formulation of Multimodal Drift Adaptation:} We present, to the best of our knowledge, the first formal integration of concept drift theory into a multimodal learning setting that combines text and image data. We define the multimodal concept drift problem and discuss the challenges that arise when the informational value of one modality changes over time. Additionally, we identify shortcomings in current fusion methods and emphasize the necessity for an adaptive fusion mechanism capable of dynamically adjusting the contributions of each modality.

\textbf{Adaptive Control Strategy for Learning:} We have developed an online adaptation controller that monitors the model's performance and dynamically adjusts two critical elements of the learning process: (1) the model's learning rate, which governs how quickly the model learns from new data compared to how well it retains information from past data, and (2) the fusion weighting between different modalities, which prioritizes the more reliable modality when drift is detected in the other. This controller is rule-based and lightweight. It detects drift through statistical changes in the model's error over time and triggers adjustments to the learning rate ($\eta$) and the fusion coefficient ($\alpha$). Our approach treats the prediction error as a feedback signal and establish adaptation rules similar to a PID controller to minimize the error. 

\textbf{Lyapunov Stability Analysis:} A key theoretical contribution of our work is developing a Lyapunov function for the adaptive learning system. We present two new theorems that establish conditions for the stability of the closed-loop learning process (comprising the model and the controller) in the sense of Lyapunov. Specifically, we demonstrate that, under mild assumptions, the system's error will remain bounded during periods of continuous drift and will converge to zero (or to an arbitrarily small value) once the drift ceases.


\section{Related Work}
\textbf{Concept Drift Detection and Adaptation:}
Drift detection techniques fall into three categories: (1) Statistical Input Monitoring: testing distribution differences between recent and earlier data using methods like Kullback-Leibler (KL) divergence~\cite{dasu2006information} or adaptive windowing (ADWIN)~\cite{liu2017fuzzy}; (2) Performance Monitoring: tracking error rates against statistical thresholds~\cite{gama2004learning, gama2006learning, baena2006early, xu2017dynamic}, effective in supervised settings; and (3) Parameter Tracking: adaptation strategies include model retraining, prioritizing recent data, or using ensembles where new learners handle emerging patterns~\cite{bach2008paired,bifet2007learning,liu2016fp}. Most methods are designed for single-modality streams and may not suit multimodal systems where drift can affect individual modalities.  

Recent advancements in deep learning have spurred research into concept drift in neural networks and large models. While continual learning and concept drift address learning from non-stationary data, continual learning typically prevents catastrophic forgetting of previous tasks. In contrast, concept drift emphasizes rapid adaptation to evolving data. Some studies, like~\cite{yang2024adapting}, propose extending concept drift theory to multimodal language models, recognizing their vulnerability to gradual shifts and sudden out-of-distribution events. During pre-training, they introduced a drift adapter module to address drift, highlighting a growing focus on this issue in complex AI systems. However, their approach primarily targets pre-training bias rather than real-time adaptation. Our work stands apart by addressing online drift during deployment while ensuring stability.

\textbf{Multimodal Fusion and Learning Techniques:}
Combining modalities improves prediction but presents fusion challenges~\cite{baltruvsaitis2018multimodal}. Methods include early fusion (feature-level merging), late fusion (decision-level integration), bilinear pooling, attention mechanisms, and cross-modal transformers~\cite{papadopoulos2025red}. In fake news detection, specialized architectures have evolved from EANN's~\cite{wang2018eann} adversarial training for topic-invariant representations to SpotFake~\cite{singhal2019spotfake} and MMDFND's~\cite{tong2024mmdfnd} attention-based approaches. Recent bidirectional cross-modal fusion (BCMF)~\cite{yu2022bcmf} enables information flow between text and image encoders, representing significant progress in multimodal integration techniques. 

Despite recent advancements, most multimodal models operate under a fixed fusion strategy once trained. During training, the model learns to weigh and utilize different modalities, but these weights remain static afterward. If a particular modality's relevance changes over time, the model cannot easily reconfigure itself due to drift. For example, a fake news detection model might initially rely heavily on textual patterns to identify deception. However, if bad actors use innocuous text paired with deceptive images over time, the model's original fusion scheme could misallocate attention and risk missing the visual cues. Some recent works~\cite{papadopoulos2025red, xue2023dynamic, han2022multimodal} have explored dynamic modulation of importance; for instance, a model might use an attention mechanism to emphasize the modality that provides a more significant "signal" for a given sample. However, this approach usually focuses on the content of the sample itself rather than detecting distributional shifts over time. In contrast, our proposed method explicitly adjusts fusion weights in response to drift over time. We conceptualize this as an adaptive late fusion approach, introducing a controllable parameter $\alpha(t)$ to interpolate between modalities. This idea of adaptive fusion has limited precedents; related concepts include gating networks or conditional fusion, in which an auxiliary network predicts fusion weights for each sample. We go further by making $\alpha$ a function of time and past errors, rather than relying on the current sample, thus directly addressing the temporal drift in modality utility. 

\section{Preliminary and Problem Formulation}\label{sec:prel}
\textbf{Concept Drift Definition:} 
We consider a supervised learning problem in an online setting. At each time step $t = 1, 2, 3, \dots$, the system receives a new sample $(x_t^{(1)}, x_t^{(2)}, \dots, x_t^{(M)},, y_t)$, where $x_t^{(M)}$ denotes the input from the $M$-th modality (for instance, $M=2$, $x^{(1)}$ is text and $x^{(2)}$ is image), and $y_t$ is the ground-truth label or target. The crucial aspect is that the data distribution is time-dependent. Let $\mathcal{D}_t$ denote the distribution generating $(x_t^{(1)},\dots,x_t^{(M)}, y_t)$ at time $t$. Concept drift means $\mathcal{D}_t$ changes over time: there exists some $t$ and $t+\Delta$ for which $\mathcal{D}_{t} \neq \mathcal{D}_{t+\Delta}$. Following standard definitions, a concept drift occurs at time $t$ if the joint probability $P_t(X^{(1)},\dots,X^{(M)}, Y)$ is statistically different from $P_{t+1}(X^{(1)},\dots,X^{(M)}, Y)$. These changes can be abrupt or gradual. This paper assumes a piecewise-constant drift for theoretical analysis (drift occurs at discrete points, dividing the timeline into relatively stable distribution segments) and later simulates the drifts.

\textbf{Multimodal Learning Model:}
Define a model $f_{\theta}(x^{(1)}, x^{(2)}, \dots, x^{(M)})$ with parameters $\theta$ that produces a prediction $\hat{y}_t$ given the multimodal input, and the model can be decomposed by modality. For instance, for $M=2$, we have two sub-networks: a text encoder $f^{(1)}(x^{(1)}; \theta_1)$ and an image encoder $f^{(2)}(x^{(2)}; \theta_2)$, producing modality-specific feature representations. A fusion function then combines these. In our design, the fusion is a weighted combination: let $z_t^{(1)}$ and $z_t^{(2)}$ be scalar logit outputs (final hidden features) from the text and image pipelines, respectively. We define the fused logit 
\begin{align}
    z_t = \alpha_t \cdot z_t^{(1)} + (1-\alpha_t) \cdot z_t^{(2)},
\end{align}
where $\alpha_t \in [0,1]$ is a fusion weight at time $t$. The prediction $\hat{y}_t$ is then obtained by applying a sigmoid or softmax to $z_t$. The parameter $\alpha_t$ effectively controls the influence of each modality: $\alpha_t = 0.5$ would weight them equally, $\alpha_t \to 1$ relies mostly on modality 1, and $\alpha_t \to 0$ relies on modality 2. The rest of $\theta$ encompasses the weights of $f^{(1)}$, $f^{(2)}$, and possibly any additional layers (e.g., if the features are concatenated and passed to an MLP, those MLP weights are in $\theta$).

\textbf{Loss and Error Measures:}
Let $L(\hat{y}_t, y_t)$ be the loss at time $t$ (e.g. cross-entropy loss for classification). We define the instantaneous prediction error $e_t$ as the difference between the predicted output and the true output in an appropriate sense. For analysis, assume $e_t$ is a real-valued error signal (for instance, $e_t$ could be the difference in probability assigned to the true class vs. the ideal, or some bounded transformation of the loss). We assume $e_t = 0$ corresponds to perfect prediction, and larger $|e_t|$ means worse performance. The exact definition of $e_t$ will be specified in proofs, but intuitively, one can think of $e_t$ as proportional to the classification error or loss at time $t$. We also define a desired equilibrium for this error: we desire $e_t \to 0$. However, under drift, the minimum achievable error might change over time as the optimal model changes.

\textbf{Adaptation Controller:}
We augment the learning process with an adaptive controller that adjusts two sets of parameters over time: the model weights $\theta$ are updated via stochastic gradient descent (or a similar optimizer) with a learning rate $\eta_t$, and the fusion weight $\alpha_t$ is also updated on a slower timescale. The controller has two primary responsibilities: drift detection and parameter adaptation.

Drift detection detects when a significant increase in error $e_t$ indicates a possible concept drift. This could be done by checking if $e_t$ exceeds some threshold or deviates significantly from a long-term error trend. In our implementation, we use a sliding window over the recent $W$ samples to compute
\begin{align}
    \bar{e}_t = \frac{1}{W}\sum_{i=t-W+1}^t e_i.
\end{align}
The controller flags a drift onset if $\bar{e}_t$ exceeds a dynamic threshold (for example, mean error + $k$ standard deviations in a reference period). Furthermore, parameter adaptation works when drift is detected, the controller computes adjustments $\Delta \eta$ and $\Delta \alpha$. Specifically, we define update rules:
\begin{align}
    \eta_{t+1} &= \eta_t + \Delta \eta_t, \\  \alpha_{t+1} &= \alpha_t + \Delta \alpha_t,
\end{align}
to reduce future errors. 

The adaptation logic is as follows: (1) Learning rate adaptation: If the error is increasing, this suggests the current model is not keeping up with changes. The controller increases the learning rate: $\Delta \eta_t$ is set to a positive value proportional to the error increase and possibly the error magnitude. A simple rule could be $\Delta \eta_t = k_{\eta} (e_t - e_{t-1})$ when $e_t > e_{t-1}$, for some gain $k_{\eta}>0$. To prevent divergence, we also cap $\eta_t$ not to exceed a safe maximum. Conversely, if error starts decreasing (model catching up), the controller may slowly decrease $\eta$ to a normal level to avoid overshooting (much like lowering gain as we approach a setpoint). This mechanism ensures fast learning during drift and stable convergence afterward. (2) Fusion weight adaptation: The controller adjusts $\alpha_t$ to put more weight on the modality that currently appears more reliable. We can estimate modality-specific error contributions. For instance, we can compute an "imagined" prediction using only modality 1, $\hat{y}_t^{(1)} = \text{sigmoid}(z_t^{(1)})$, and similarly $\hat{y}_t^{(2)}$ using only modality 2. By comparing these to $y_t$, the controller can judge which modality alone would perform better on recent data. Suppose modality 1's predictions are significantly more accurate than modality 2's. In that case, it implies modality 1 carries the useful signal and modality 2 might be drifting or noisy, so we increase $\alpha$ to rely more on modality 1. In the opposite case, we decrease $\alpha$. Formally, one could maintain separate error metrics $e_t^{(1)}$ and $e_t^{(2)}$ for the two modalities. Then, $\Delta \alpha_t$ can be set by a rule like $\Delta \alpha_t = k{\alpha}\big(e_t^{(2)} - e_t^{(1)}\big)$: if modality 2’s error is higher, $\Delta \alpha_t$ is positive (increase weight on modality 1); if modality 1 error is higher, $\Delta \alpha_t$ is negative (shift weight to modality 2). We clamp $\alpha_t$ within [0,1]. This update might be done conservatively (lower frequency or smaller step size) than the learning rate, since fusion changes too often could destabilize training. This slow actuator nudges the model's attention towards the currently best modality.

\textbf{State-Space Representation:}
We can conceptualize the system's evolution with a state vector that includes at least the error and possibly the parameter deviations. For analysis, consider the state as $x_t = e_t$ (the error at time $t$). The error evolves depending on the changing data distribution and our adaptive updates. We update model weights online using mini-batch stochastic gradients. A simplified dynamics can be written as:
\begin{align}
    e_{t+1} = F(e_t, \alpha_t, \eta_t,; \delta_t),
\end{align}
where $\delta_t$ represents the disturbance due to concept drift (i.e., how the optimal model's output would change from $t$ to $t+1$ due to distribution shift). The exact form of $F$ is complex since it involves the data and the model's optimization landscape. However, we can reason qualitatively: if there were no drift ($\delta_t=0$) and the model perfectly matched the current concept, then ideally $e_t$ would be zero for all $t$. If the model is imperfect but we fix $\alpha$ and a small $\eta$, gradient descent will tend to reduce $e_t$ over time toward some residual error (training convergence). Under drift ($\delta_t \neq 0$), the error will increase unless $\eta_t$ is large enough to track the change. Our controller effectively modulates $F$ by changing $\eta_t$ and $\alpha_t$.

Furthermore, we introduce a candidate Lyapunov function to analyze stability:
\begin{align}
    V(t) = \frac{1}{2} e_t^2.
\end{align}
This $V$ is always nonnegative and $V=0$ only when the error is zero, meeting the criteria of a Lyapunov function candidate. We aim to design the adaptation laws ($\Delta \eta, \Delta \alpha$ as functions of $e$ and possibly $\Delta e$) such that $V(t)$ does not increase significantly and preferably decreases after some transient, despite drift. 

\begin{figure}[!htbp]
    \centering
    \includegraphics[scale=0.4]{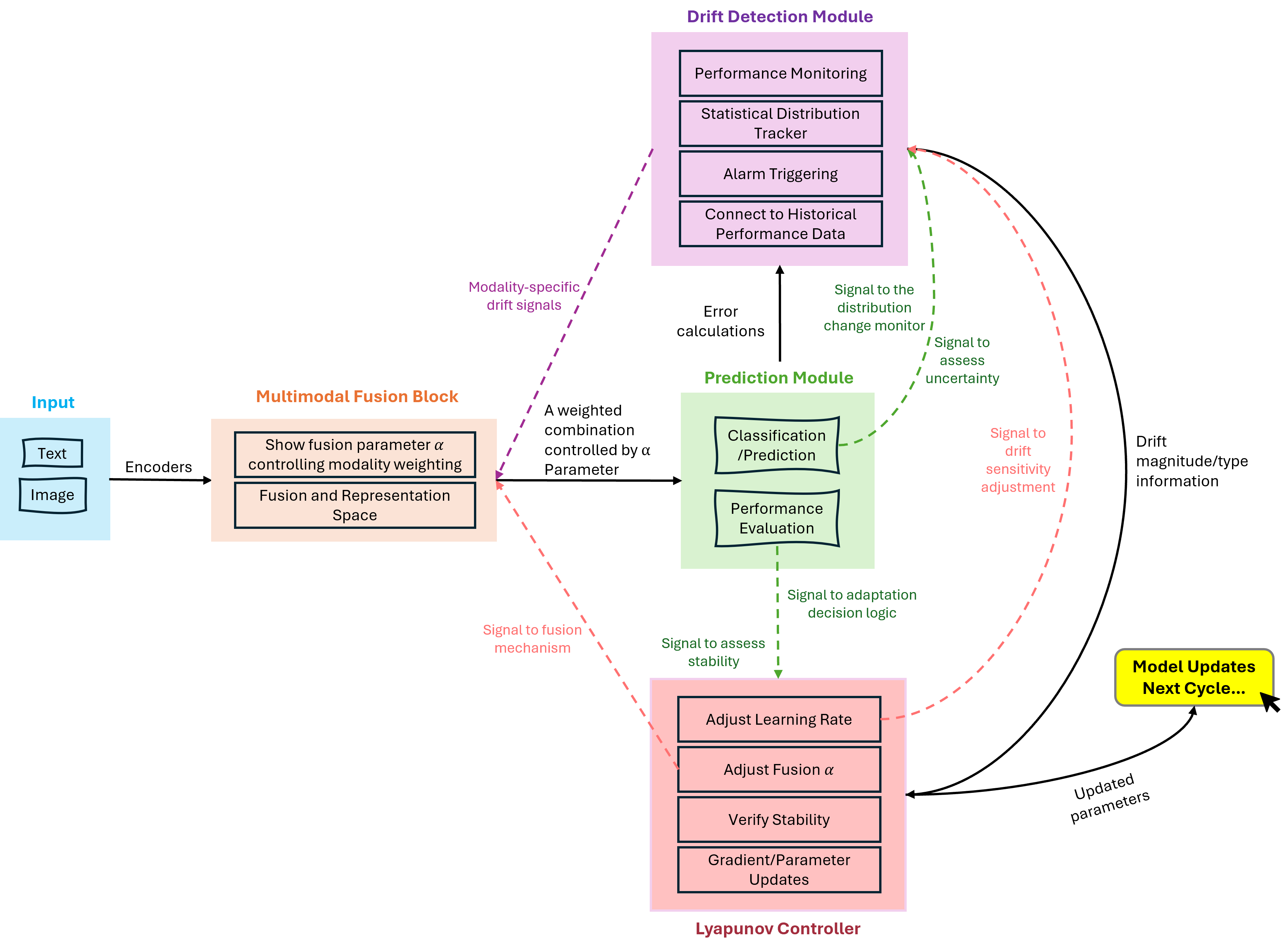}
    \caption{Illustrations of the proposed system architecture. Dashed and solid arrows represent the feedback signals and main data flow, respectively.}
    \label{fig:lsogd}
\end{figure}

\textbf{Proposed Algorithm:}
Algorithms~\ref{alg:ls_ogd_main_loop} and~\ref{alg:ls_ogd_controller_logic} in Appendix~\ref{app:algo} outline the main online learning process and the controller logic. Figure~\ref{fig:lsogd} also presents the proposed system architecture.  

\section{Main Results}\label{sec:res}
This section presents the theoretical analyses of the proposed LS-ODG framework. Our analyses establish the learning system's stability and adaptability when operating in non-stationary multimodal environments. 

\subsection{Foundational Axioms and System Properties}
To analyze the main framework, three axioms are defined:

\textbf{Axiom 1 (Bounded Drift Rate)} Concept drift, represented by the perturbation $\delta_t$ in the error dynamics or by the rate of change of optimal model parameters $||\theta^*_{t+1}-\theta^*_t||$, is assumed to be bounded. That is, there exists a constant $\delta_{max}>0$ such that $||\delta_t||\leq \delta_{max}$ for all $t$. 

This axiom is fundamental because an arbitrarily fast or large drift would make it impossible for any adaptive system with finite learning capacity to maintain low error. It ensures that the ``target" the system is trying to track does not move so erratically as to preclude effective adaptation. 

\textbf{Axiom 2 (Sufficient Model Capacity)} The multimodal learning model $f_{\theta}$ possesses sufficient capacity such that for any stationary concept $\mathcal{D}$ encountered (i.e., a period where $\mathcal{D}_t$ is fixed), there exist optimal parameters $\theta^*$ and a fusion weight $\alpha^*$ that can achieve the minimum error $e_t(\theta^*,\alpha^*)$. 

This axiom ensures that performance limitations are primarily due to drift or adaptation transients, not due to the model's inherent inability to learn the underlying concepts. It allows us to focus the analysis on the dynamics of adaptation rather than on model approximation errors for a fixed concept. 

\textbf{Axiom 3 (Lipschitz-like Error Dynamics)} The error dynamics function $F(e_t, \alpha_t, \eta_t; \delta_t)$ and the loss landscape exhibit local Lipschitz continuity with respect to $e_t$ and the adaptable parameters $\theta$ (implicitly via $\eta_t$) and $\alpha_t$. This means that small changes in parameters or error lead to proportionally bounded changes in the subsequent error or its gradient. 

This axiom ensures that the gradient-based updates and the adaptive adjustments lead to predictable changes in the error, preventing chaotic or overly sensitive responses. This helps analyze the system using local approximations and ensures that the Lyapunov function's evolution is well-behaved.

\subsection{Characterizing the Evolution of the Lyapunov Function}
Before presenting the main stability theorem, we introduce Lemma 1. This Lemma characterizes the change in the Lyapunov function $\Delta V(t) = V(t+1) - V(t)$ over one time step, under the influence of the learning process, the adaptive controller, and the external drift. 

\textbf{Lemma 1 (Bounded Increment of the Lyapunov Function with Adaptation)}
Given the system dynamics $e_{t+1}=F(e_t, \alpha_t, \eta_t; \delta_t)$ and the adaptive update rules for $\eta_t$ and $\alpha_t$, as described in Section~\ref{sec:prel} and Algorithm~\ref{alg:ls_ogd_controller_logic}. Assuming Axiom 1-3 hold, and the learning rate $\eta_t$ is positive and the adaption gains $k_{\eta}$ and $k_{\alpha}$ are chosen within stable regions, then the change in the Lyapunov function, $\Delta V(t) = V(t+1) - V(t)$, can be bounded as:
\begin{align}
    \Delta V(t) \leq-\gamma_1 \eta_t e_t^2+\gamma_2 \eta_t\left|e_t\right|\left\|\delta_t\right\|+\gamma_3 \eta_t^2\left(\mathcal{G}\left(e_t, \alpha_t\right)+\left\|\delta_t\right\|^2\right)+\mathcal{O}\left(\left\|\Delta \alpha_t\right\|\right),
\end{align}
where $\gamma_1$, $\gamma_2$ and $\gamma_3$ are positive constants. 

In Lemma 1, the first term $- \tilde{\gamma_1}\eta_te^2_t$ is referred to as the stabilizing term and reflects the error reduction due to the learning algorithm attempting to minimize the current loss. $\gamma_1$ is related to the learning effectiveness, i.e., a local convexity or gradient dominance property from Axiom 3. The second term $\gamma_2\eta_t |e_t| ||\delta_t||$ is referred to as the disturbance term, which measures the impact of concept drift on the error. The third term $\gamma_3 \eta_t^2\left (\mathcal{G}\left(e_t, \alpha_t\right)+\left\|\delta_t\right\|^2\right)$ is the noise term, evaluating the nonlinearities or gradient stochasticity in SGD. The final term $\mathcal{O}\left(\left\|\Delta \alpha_t\right\|\right)$ is referred to as the fusion adaptation term, representing the effect of changing fusion weight on the error.

For the candidate Lyapunov function $V(t) = \frac{1}{2} e_t^2$, Lemma 1 establishes that the stabilizing effect of learning must overcome all three other terms to decrease error. This balance is mediated by the adaptive learning rate $\eta_t$. For instance, if $|e_t|$ is large, the controller tends to increase $\eta_t$, therefore increasing the stabilizing term. Additionally, since $\Delta V(t)$ remains negative when $|e_t|$ is sufficiently large, even when $\delta_t \neq 0$, Lemma 1 forms the basis for proving the uniform ultimate boundedness (UUB). The asymptotic stability is achieved without concept drift (i.e., $\delta_t = 0$).

\textbf{Remark 1 (On the Choice of Lyapunov Function and Adaptation Gains)} 
The choice of $V(t)=\frac{1}{2} e_t^2$ is an intuitive selection for tracking error, relating the ``energy" of the system to the squared prediction error. The condition in Lemma 1 and Theorem 1 regarding ``appropriately small" adaptation gains ($k_\eta, k_\alpha$) is common in adaptive control. If gains are too large, the system might overreact to errors or drift, leading to oscillations or instability, violating the conditions needed for $\Delta V(t)$ to be reliably negative or bounded as desired. The theoretical existence of such stable gain regions is key, and practical selection often involves tuning guided by these theoretical constraints.

\subsection{System Stability and Error Convergence under Bounded Concept Drift}
The first main theoretical result concerning the overall stability of the adaptive learning system is presented with the Axioms and Lemma 1.

\textbf{Theorem 1 (Lyapunov Stability under Bounded Drift)} Consider an adaptive multimodal learning system with error dynamics characterized by $e_{t+1}=$ $F\left(e_t, \alpha_t, \eta_t; \delta_t\right)$, operating under Axioms 1-3. 

Assume that the adaptive controller adjusts $\eta_t$ and $\alpha_t$ such that Lemma 1 holds, and the adaptation gains $k_\eta$ and $k_\alpha$ are appropriately small, then the closed-loop learning system is Lyapunov stable (i.e., the prediction error $e_t$ is UUB). This implies that the ultimate bound on the error $\varepsilon_B$ and the bound on the initial error $B_0$ exist, and are positive constants, such that the error trajectory would remain bounded, and eventually stays within the ultimate bound $\varepsilon_B$ if $\left|e_{t_0}\right|<B_0$, where $e_{t_0}$ denotes the initial error. 


Theorem 1 establishes a formal Lyapunov stability guarantee in adaptive multimodal learning under concept drift for the first time. Unlike heuristic approaches, this provides a mathematical foundation for the system's behavior. The UUB acknowledges that the error may not always converge to zero in persistent drift. Instead, UUB guarantees that the error will eventually enter and remain within a small, predictable region around zero. The size of this region, $\varepsilon_B$, will depend on the magnitude of the drift (Axiom 1) and the system's capacity to adapt. Also, this theorem guarantees that if the environment stabilizes (i.e., $\delta_t \rightarrow 0$), the system will fully recover, with the prediction error converging to zero. This demonstrates the system's ability to not only manage ongoing drift but also achieve optimal performance once stability is restored. 

\textbf{Corollary 1 (Asymptotic Stability in Stationary Environments)} 
If the environment is stationary from the outset (i.e., $\delta_t=0$ for all $t \geq t_0$), and Axioms 2-3 hold, then the LS-OGD system with an appropriately chosen, potentially constant, learning rate $\eta$ and fusion weight $\alpha$ ensures that the prediction error $e_t$ converges asymptotically to zero (i.e., $e_t \rightarrow 0$ as $t \rightarrow \infty$).

This corollary simplifies Theorem 1 for the non-drifting case. It confirms that the learning mechanism within LS-OGD is sound and can achieve optimal performance when the environment is stable. The UUB property of Theorem 1 then extends this, showing robustness by guaranteeing bounded error even when the stationarity assumption is violated by bounded drift.

\subsection{Targeted Adaptation of the Fusion Weight}
The second main theorem concerns the system's ability to adapt its fusion strategy. This relies on the controller's ability to adjust $\alpha_t$ correctly. Hence, Lemma 2 is presented as a key for Theorem 2.

\textbf{Lemma 2 (Convergent Dynamics of Fusion Weight $\alpha_t$ under Unilateral Modality Drift)}
Given the fusion weight $\alpha_t$ on modality 1, the fusion weight on modality 2 is defined as $1-\alpha_t$. 

Consider the fusion weight update rule:
\begin{align}
    \alpha_{t+1}=\operatorname{clip}_{[0,1]}\left(\alpha_t+\Delta \alpha_t\right), \quad \Delta \alpha_t= k_{\alpha}\left(e_t^{(2), \text{est}} - e_t^{(1), \text{est}}\right)
\end{align}
where $e_t^{(m), \text{est}}$ being the estimated error from modality $m$, and $k_\alpha>0$ is the adaptation gain.

Following the update rule for fusion weight, if modality 1 becomes less reliable than modality 2, leading to $e_t^{(1), \text{est}}>e_t^{(2), \text{est}}+\epsilon_\alpha$ for $t>t_0$ for some positive margin $\epsilon_{\alpha}$, then: 
\begin{enumerate}[(i)]
    \item $\alpha_t$ will tend to decrease.
    \item Provided $k_{\alpha}$ is sufficiently small to ensure smooth adaptation and prevent overshoot. The error estimation difference is persistent, $\alpha_t$ will converge towards 0 or a small value close to 0 if the error difference is not consistently large or noise is present in the error estimates. 
\end{enumerate}
Lemma 2 describes how the controller adjusts $\alpha_t$ as one of the modalities is less reliable. The update rule is a form of gradient descent on an implicit objective that seeks to minimize reliance on the modality with higher perceived error. It confirms that the fusion adaptation mechanism reduces the weight of a consistently underperforming modality, even being able to
``switch of'' a compromised modality by setting $\alpha_t$ to 0 or 1 depending on which modality is less reliable.  


\textbf{Proposition 1 (Convergence Rate of Fusion Weight $\alpha_t$)}
Under the conditions of Lemma 2, assume that modality 1 is consistently with low reliability (i.e., $e_t^{(1), \text{est}}>e_t^{(2), \text{est}}+\epsilon_\alpha$ for all $t>t_0$), the fusion weight is expected to be 0. Given that $\alpha_t$ is not yet converged to its boundary, then the expected decreasing rate of $\alpha_t$ is approximately linear: 
\begin{align}
    \mathbb{E}\left[\alpha_{t+1}-\alpha_t\right] \approx-k_\alpha\left(\mathbb{E}\left[e_t^{(1), \text{est}}\right]-\mathbb{E}\left[e_t^{(2), \text{est}}\right]\right)
\end{align}
The steps to boundary convergence is therefore inversely proportional to $k_\alpha$  and the persistent magnitude of the estimated error difference, barring noise in error estimation.


Proposition 1 provides insight into the dynamics of the fusion adaptation. While Lemma 2 states the tendency and convergence, this suggests that the speed of this adaptation is tunable via $k_{\alpha}$ and directly influenced by how clearly the system can distinguish the performance of the modalities. A larger error difference allows for faster isolation of the faulty modality. This has a practical implication for setting $k_{\alpha}$: a larger $k_{\alpha}$ leads to faster fusion adaptation but must be balanced against the risk of instability or overreaction to noisy error estimates (as per Remark 1). Building upon Lemma 2 and Proposition 1, Theorem 2 establishes the system's robust response in the case of severe drift affecting a single modality. 

\textbf{Theorem 2 (Modality Adaptation and Error Reduction under Severe Unilateral Drift)} 
Consider the same scenario defined previously, where modality 1 is consistently less reliable. If the adaptive controller adjusts $\alpha_t$ such that Lemma 2 holds, then:
\begin{enumerate}[(i)]
    \item The fusion weight $\alpha_t$ will converge towards 0, decreasing and eventually nullifying the influence of the compromised modality 1. 
    \item Consequently, the system's overall prediction error $e_t$ will, after this reconfiguration transient, asymptotically approach the error level achievable by an optimal unimodal model utilizing only the reliable modality 2. 
    \item The Lyapunov stability of the system is preserved throughout this fusion reconfiguration process and thereafter, with the ultimate error bound being determined by the performance achievable with modality 2 and any residual drift or noise in its stream. 
\end{enumerate}

Theorem 2 indicates that the internal structure of the model is actively changing, making this model different from traditional multimodal systems, as traditional systems often suffer from a persistent drop in performance~\cite{zhao2024survey,wu2024deep}.
Theorem 2 also shows that the adaptive fusion mechanism can correctly identify and isolate a ``fault'' modality by driving its weight ($\alpha_t$) towards zero. This theorem quantifies the system's performance post-adaptation as it converges to the best possible performance using the remaining valid information sources. Additionally, system behavior with partial modality degradation and practical estimation of modality-specific errors are presented in Appendix~\ref{app:results}. The full proofs for our main results are in Appendix~\ref{theoremproof1}.

\section{Experiment}
\vspace{-1mm}
To empirically validate LS-OGD's theoretical contributions and its capacity to address key challenges in multimodal concept drift, we conducted experiments on the M3A~\cite{xu2024m3a} classification dataset. After initial training in a stable environment (Phase 1), the multimodal system was subjected to significant concept drift (Phase 2). Full experimental setup~\footnote{The code is available at \href{https://github.com/Bellleuang1022/Lyapunov-Stable-Adaptive-Control-for-Multimodal-Concept-Drift.git}{https://github.com/Bellleuang1022/Lyapunov-Stable-Adaptive-Control-for-Multimodal-Concept-Drift.git}.} and additional results are detailed in Appendix~\ref{app:exp}.

First, Figure~\ref{fig:error_sig} demonstrates LS-OGD's ability to maintain stability. The controller's estimated drift signal, an empirical proxy for the bounded drift $\delta_t$ (Axiom 1), frequently activates during Phase 2. Correspondingly, while fluctuating with drift, the system's prediction error remains contained and is effectively driven down by the adaptive responses. This behavior empirically corroborates the UUB of the error, guaranteed by Theorem 1, and the error-reducing system dynamics characterized by Lemma 1, indicating stable learning in a non-stationary environment.  


\vspace{-1mm}
\begin{figure}[!htbp]
    \centering
    \includegraphics[scale=0.05]{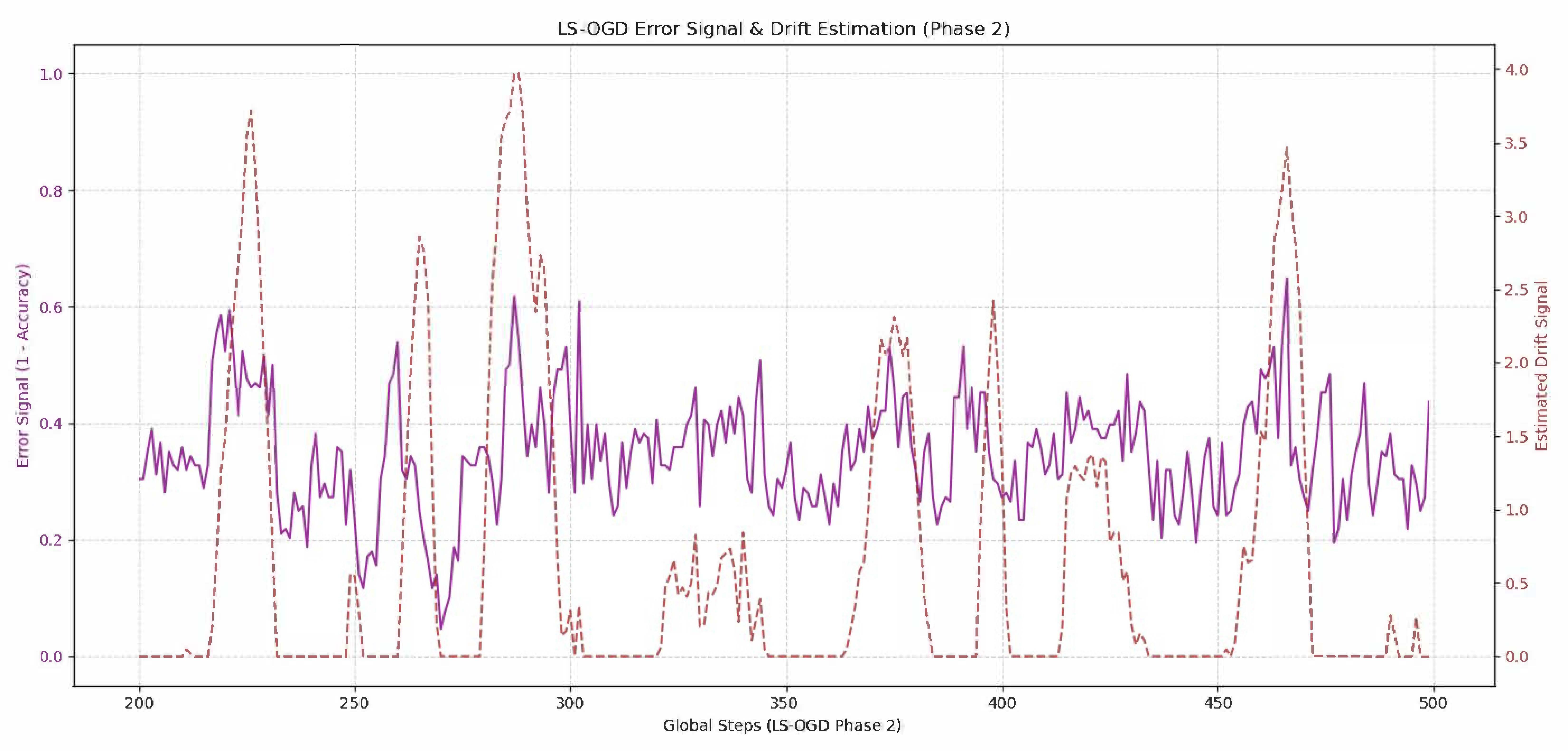}
    \caption{LS-OGD Error Signal \& Drift Estimation in Phase 2}
    \label{fig:error_sig}
\end{figure}

\begin{figure}[!htbp]
    \centering
    \includegraphics[scale=0.29]{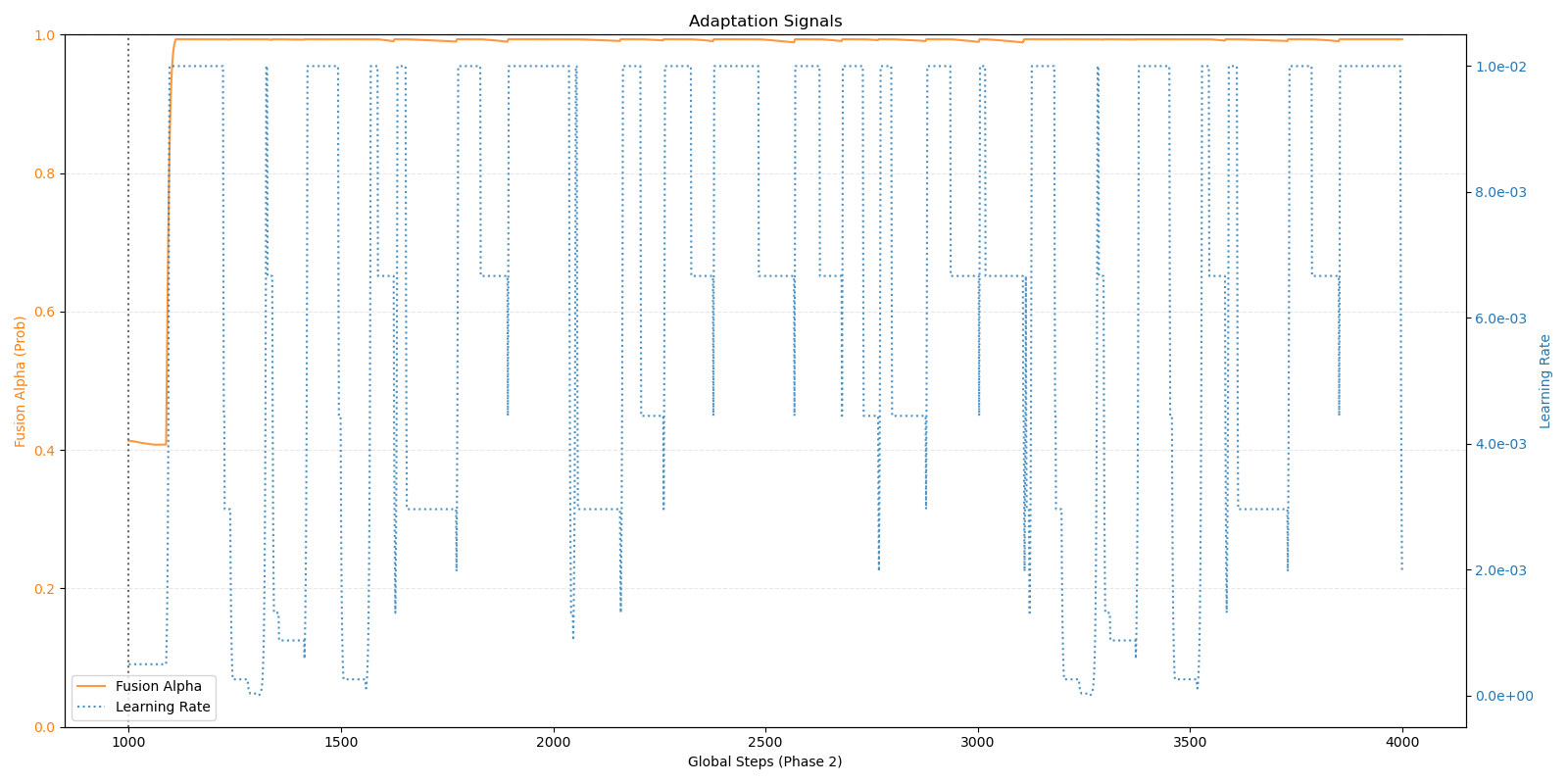}
    \caption{LS-OGD Controller Adaptation Signals in Phase 2}
    \label{fig:adap_sig}
\end{figure}

Second, Figure~\ref{fig:adap_sig} reveals how LS-OGD addresses modality-specific challenges. Dynamic learning rate adjustments enhanced plasticity during instability, aligning with Lemma 1 and stable gain principles (Remark 1). More importantly, when confronted with severe unilateral modality drift, the fusion parameter $\alpha_t$ underwent a rapid and significant shift. This sharp re-weighting, isolating the compromised modality and preserving performance by relying on the healthier one, empirically confirms the adaptive fusion dynamics (Lemma 2) and the fault-tolerant capabilities (Theorem 2). This tackles the limitations of fixed fusion strategies and demonstrates robust, stable adaptation to modality-specific drift, a core challenge highlighted in the introduction. 


\vspace{-4mm}
\section{Conclusion}

The LS-OGD framework represents a significant advancement in adaptive AI by combining Lyapunov-based control theory with online gradient learning and dynamic fusion weights. This integration allows for stability and resilience in the face of non-stationary, multimodal inputs. The framework's core assurance is that prediction errors will remain consistently bounded under limited drift and converge to zero as disturbances diminish. This ensures predictable performance, even when data distributions change. Moreover, LS-OGD features a modality-aware adaptation mechanism that automatically down-weights unreliable sensors or data streams, allowing the system to concentrate learning on trustworthy information sources. This lightweight, online approach is suited for real-world applications, such as autonomous vehicles that need to adjust the importance of inputs from vision, LiDAR, and radar under varying conditions; intelligence systems that integrate imagery, signals, and human reports while facing adversarial pressures; and misinformation detectors that combat evolving text and image manipulations. By establishing a robust framework for dynamic sensor re-weighting and bounded-error learning, LS-OGD lays a solid foundation for trustworthy AI. Its modular design also allows for extensions to nonlinear models, deep architectures, and context-aware adaptation strategies. Limitation and broader impact are discussed in Appendix~\ref{app:lim}.

\newpage
\section{Acknowledgment}
The authors sincerely thank the program and area chairs, along with the anonymous reviewers, for their constructive comments on this work. Additionally, Bell expresses her gratitude to her beloved pets, Kiwi (an Australian Shepherd) and Meemeow Pan (a Siamese Tabby), for their emotional support.

\bibliography{reference}

\begin{thebibliography}{43}
\providecommand{\natexlab}[1]{#1}
\providecommand{\url}[1]{\texttt{#1}}
\expandafter\ifx\csname urlstyle\endcsname\relax
  \providecommand{\doi}[1]{doi: #1}\else
  \providecommand{\doi}{doi: \begingroup \urlstyle{rm}\Url}\fi

\bibitem[Bach and Maloof(2008)]{bach2008paired}
Stephen~H Bach and Marcus~A Maloof.
\newblock Paired learners for concept drift.
\newblock In \emph{2008 Eighth IEEE International Conference on Data Mining}, pages 23--32. IEEE, 2008.

\bibitem[Baena-Garc{\i}a et~al.(2006)Baena-Garc{\i}a, del Campo-{\'A}vila, Fidalgo, Bifet, Gavalda, and Morales-Bueno]{baena2006early}
Manuel Baena-Garc{\i}a, Jos{\'e} del Campo-{\'A}vila, Raul Fidalgo, Albert Bifet, Ricard Gavalda, and Rafael Morales-Bueno.
\newblock Early drift detection method.
\newblock In \emph{Fourth international workshop on knowledge discovery from data streams}, volume~6, pages 77--86. Citeseer, 2006.

\bibitem[Baltru{\v{s}}aitis et~al.(2018)Baltru{\v{s}}aitis, Ahuja, and Morency]{baltruvsaitis2018multimodal}
Tadas Baltru{\v{s}}aitis, Chaitanya Ahuja, and Louis-Philippe Morency.
\newblock Multimodal machine learning: A survey and taxonomy.
\newblock \emph{IEEE transactions on pattern analysis and machine intelligence}, 41\penalty0 (2):\penalty0 423--443, 2018.

\bibitem[Bifet and Gavalda(2007)]{bifet2007learning}
Albert Bifet and Ricard Gavalda.
\newblock Learning from time-changing data with adaptive windowing.
\newblock In \emph{Proceedings of the 2007 SIAM international conference on data mining}, pages 443--448. SIAM, 2007.

\bibitem[Canonaco et~al.(2021)Canonaco, Bergamasco, Mongelluzzo, and Roveri]{canonaco2021adaptive}
Giuseppe Canonaco, Alex Bergamasco, Alessio Mongelluzzo, and Manuel Roveri.
\newblock Adaptive federated learning in presence of concept drift.
\newblock In \emph{2021 International Joint Conference on Neural Networks (IJCNN)}, pages 1--7. IEEE, 2021.

\bibitem[Chu et~al.(2024)Chu, Wei, Liu, Zhao, and Miyatake]{chu2024lyapunov}
Haoyu Chu, Shikui Wei, Ting Liu, Yao Zhao, and Yuto Miyatake.
\newblock Lyapunov-stable deep equilibrium models.
\newblock In \emph{Proceedings of the AAAI Conference on Artificial Intelligence}, volume~38, pages 11615--11623, 2024.

\bibitem[Dai et~al.(2021)Dai, Landry, Yang, Pavone, and Tedrake]{dai2021lyapunov}
Hongkai Dai, Benoit Landry, Lujie Yang, Marco Pavone, and Russ Tedrake.
\newblock Lyapunov-stable neural-network control.
\newblock \emph{arXiv preprint arXiv:2109.14152}, 2021.

\bibitem[Dasu et~al.(2006)Dasu, Krishnan, Venkatasubramanian, and Yi]{dasu2006information}
Tamraparni Dasu, Shankar Krishnan, Suresh Venkatasubramanian, and Ke~Yi.
\newblock An information-theoretic approach to detecting changes in multi-dimensional data streams.
\newblock In \emph{Proc. Symposium on the Interface of Statistics, Computing Science, and Applications (Interface)}, 2006.

\bibitem[Doicin et~al.(2016)Doicin, Popescu, and Patrascioiu]{doicin2016pid}
Bogdan Doicin, Marian Popescu, and Cristian Patrascioiu.
\newblock Pid controller optimal tuning.
\newblock In \emph{2016 8th International Conference on Electronics, Computers and Artificial Intelligence (ECAI)}, pages 1--4. IEEE, 2016.

\bibitem[Dontoh et~al.(2025)Dontoh, Ivey, Sirbaugh, Danyo, and Aboah]{dontoh2025visual}
Anthony Dontoh, Stephanie Ivey, Logan Sirbaugh, Andrews Danyo, and Armstrong Aboah.
\newblock Visual dominance and emerging multimodal approaches in distracted driving detection: A review of machine learning techniques.
\newblock \emph{arXiv preprint arXiv:2505.01973}, 2025.

\bibitem[Gama and Castillo(2006)]{gama2006learning}
Joao Gama and Gladys Castillo.
\newblock Learning with local drift detection.
\newblock In \emph{International conference on advanced data mining and applications}, pages 42--55. Springer, 2006.

\bibitem[Gama et~al.(2004)Gama, Medas, Castillo, and Rodrigues]{gama2004learning}
Joao Gama, Pedro Medas, Gladys Castillo, and Pedro Rodrigues.
\newblock Learning with drift detection.
\newblock In \emph{Advances in Artificial Intelligence--SBIA 2004: 17th Brazilian Symposium on Artificial Intelligence, Sao Luis, Maranhao, Brazil, September 29-Ocotber 1, 2004. Proceedings 17}, pages 286--295. Springer, 2004.

\bibitem[Gong et~al.(2022)Gong, Jeong, Kim, Kim, Shin, and Lee]{gong2022note}
Taesik Gong, Jongheon Jeong, Taewon Kim, Yewon Kim, Jinwoo Shin, and Sung-Ju Lee.
\newblock Note: Robust continual test-time adaptation against temporal correlation.
\newblock \emph{Advances in Neural Information Processing Systems}, 35:\penalty0 27253--27266, 2022.

\bibitem[Gong et~al.(2023)Gong, Kim, Lee, Chottananurak, and Lee]{gong2023sotta}
Taesik Gong, Yewon Kim, Taeckyung Lee, Sorn Chottananurak, and Sung-Ju Lee.
\newblock Sotta: Robust test-time adaptation on noisy data streams.
\newblock \emph{Advances in Neural Information Processing Systems}, 36:\penalty0 14070--14093, 2023.

\bibitem[Han et~al.(2022)Han, Yang, Huang, Zhang, and Yao]{han2022multimodal}
Zongbo Han, Fan Yang, Junzhou Huang, Changqing Zhang, and Jianhua Yao.
\newblock Multimodal dynamics: Dynamical fusion for trustworthy multimodal classification.
\newblock In \emph{Proceedings of the IEEE/CVF conference on computer vision and pattern recognition}, pages 20707--20717, 2022.

\bibitem[Janakiraman et~al.(2013)Janakiraman, Nguyen, and Assanis]{janakiraman2013lyapunov}
Vijay~Manikandan Janakiraman, XuanLong Nguyen, and Dennis Assanis.
\newblock A lyapunov based stable online learning algorithm for nonlinear dynamical systems using extreme learning machines.
\newblock In \emph{The 2013 International Joint Conference on Neural Networks (IJCNN)}, pages 1--8. IEEE, 2013.

\bibitem[Li et~al.(2024)Li, Chen, Rao, Guan, Jiang, and Li]{li2024multimodal}
Kendi Li, Di~Chen, Zuguang Rao, Zijing Guan, Ya~Jiang, and Yuanqing Li.
\newblock A multimodal asynchronous human-machine interaction method based on electrooculography and speech recognition for wheelchair control.
\newblock \emph{IEEE Sensors Journal}, 2024.

\bibitem[Liu et~al.(2017{\natexlab{a}})Liu, Song, Zhang, and Lu]{liu2017regional}
Anjin Liu, Yiliao Song, Guangquan Zhang, and Jie Lu.
\newblock Regional concept drift detection and density synchronized drift adaptation.
\newblock In \emph{IJCAI international joint conference on artificial intelligence}, 2017{\natexlab{a}}.

\bibitem[Liu et~al.(2017{\natexlab{b}})Liu, Zhang, and Lu]{liu2017fuzzy}
Anjin Liu, Guangquan Zhang, and Jie Lu.
\newblock Fuzzy time windowing for gradual concept drift adaptation.
\newblock In \emph{2017 IEEE international conference on fuzzy systems (FUZZ-IEEE)}, pages 1--6. IEEE, 2017{\natexlab{b}}.

\bibitem[Liu et~al.(2016)Liu, Wu, and Jiang]{liu2016fp}
Dong Liu, YouXi Wu, and He~Jiang.
\newblock Fp-elm: An online sequential learning algorithm for dealing with concept drift.
\newblock \emph{Neurocomputing}, 207:\penalty0 322--334, 2016.

\bibitem[Liu et~al.(2025)Liu, Yan, Liu, Fu, Wen, Liu, and Li]{liu2025exploring}
Moyang Liu, Kaiying Yan, Yukun Liu, Ruibo Fu, Zhengqi Wen, Xuefei Liu, and Chenxing Li.
\newblock Exploring modality disruption in multimodal fake news detection.
\newblock \emph{arXiv preprint arXiv:2504.09154}, 2025.

\bibitem[Ma et~al.(2022)Ma, Vemprala, Wang, Gupta, Song, McDufft, and Kapoor]{ma2022compass}
Shuang Ma, Sai Vemprala, Wenshan Wang, Jayesh~K Gupta, Yale Song, Daniel McDufft, and Ashish Kapoor.
\newblock Compass: Contrastive multimodal pretraining for autonomous systems.
\newblock In \emph{2022 IEEE/RSJ International Conference on Intelligent Robots and Systems (IROS)}, pages 1000--1007. IEEE, 2022.

\bibitem[Miyaguchi and Kajino(2019)]{miyaguchi2019cogra}
Kohei Miyaguchi and Hiroshi Kajino.
\newblock Cogra: Concept-drift-aware stochastic gradient descent for time-series forecasting.
\newblock In \emph{Proceedings of the AAAI Conference on Artificial Intelligence}, volume~33, pages 4594--4601, 2019.

\bibitem[Niu et~al.(2023)Niu, Wu, Zhang, Wen, Chen, Zhao, and Tan]{niu2023towards}
Shuaicheng Niu, Jiaxiang Wu, Yifan Zhang, Zhiquan Wen, Yaofo Chen, Peilin Zhao, and Mingkui Tan.
\newblock Towards stable test-time adaptation in dynamic wild world.
\newblock \emph{arXiv preprint arXiv:2302.12400}, 2023.

\bibitem[Papadopoulos et~al.(2025)Papadopoulos, Koutlis, Papadopoulos, and Petrantonakis]{papadopoulos2025red}
Stefanos-Iordanis Papadopoulos, Christos Koutlis, Symeon Papadopoulos, and Panagiotis~C Petrantonakis.
\newblock Red-dot: Multimodal fact-checking via relevant evidence detection.
\newblock \emph{IEEE Transactions on Computational Social Systems}, 2025.

\bibitem[Patil et~al.(2024)Patil, Le, Griffis, and Dixon]{patil2024lyapunov}
Omkar~Sudhir Patil, Duc~M Le, Emily~J Griffis, and Warren~E Dixon.
\newblock Lyapunov-based deep residual neural network (resnet) adaptive control.
\newblock \emph{arXiv preprint arXiv:2404.07385}, 2024.

\bibitem[Singhal et~al.(2019)Singhal, Shah, Chakraborty, Kumaraguru, and Satoh]{singhal2019spotfake}
Shivangi Singhal, Rajiv~Ratn Shah, Tanmoy Chakraborty, Ponnurangam Kumaraguru, and Shin'ichi Satoh.
\newblock Spotfake: A multi-modal framework for fake news detection.
\newblock In \emph{2019 IEEE fifth international conference on multimedia big data (BigMM)}, pages 39--47. IEEE, 2019.

\bibitem[Song et~al.(2023)Song, Lee, Kweon, and Choi]{song2023ecotta}
Junha Song, Jungsoo Lee, In~So Kweon, and Sungha Choi.
\newblock Ecotta: Memory-efficient continual test-time adaptation via self-distilled regularization.
\newblock In \emph{Proceedings of the IEEE/CVF Conference on Computer Vision and Pattern Recognition}, pages 11920--11929, 2023.

\bibitem[Sun et~al.(2021)Sun, Greene, Le, Bell, Chowdhary, and Dixon]{sun2021lyapunov}
Runhan Sun, Max~L Greene, Duc~M Le, Zachary~I Bell, Girish Chowdhary, and Warren~E Dixon.
\newblock Lyapunov-based real-time and iterative adjustment of deep neural networks.
\newblock \emph{IEEE Control Systems Letters}, 6:\penalty0 193--198, 2021.

\bibitem[Tong et~al.(2024)Tong, Lu, Zhao, Lai, and Shi]{tong2024mmdfnd}
Yu~Tong, Weihai Lu, Zhe Zhao, Song Lai, and Tong Shi.
\newblock Mmdfnd: Multi-modal multi-domain fake news detection.
\newblock In \emph{Proceedings of the 32nd ACM International Conference on Multimedia}, pages 1178--1186, 2024.

\bibitem[Wang et~al.(2020)Wang, Shelhamer, Liu, Olshausen, and Darrell]{wang2020tent}
Dequan Wang, Evan Shelhamer, Shaoteng Liu, Bruno Olshausen, and Trevor Darrell.
\newblock Tent: Fully test-time adaptation by entropy minimization.
\newblock \emph{arXiv preprint arXiv:2006.10726}, 2020.

\bibitem[Wang et~al.(2022)Wang, Fink, Van~Gool, and Dai]{wang2022continual}
Qin Wang, Olga Fink, Luc Van~Gool, and Dengxin Dai.
\newblock Continual test-time domain adaptation.
\newblock In \emph{Proceedings of the IEEE/CVF Conference on Computer Vision and Pattern Recognition}, pages 7201--7211, 2022.

\bibitem[Wang et~al.(2018)Wang, Ma, Jin, Yuan, Xun, Jha, Su, and Gao]{wang2018eann}
Yaqing Wang, Fenglong Ma, Zhiwei Jin, Ye~Yuan, Guangxu Xun, Kishlay Jha, Lu~Su, and Jing Gao.
\newblock Eann: Event adversarial neural networks for multi-modal fake news detection.
\newblock In \emph{Proceedings of the 24th acm sigkdd international conference on knowledge discovery \& data mining}, pages 849--857, 2018.

\bibitem[Wu et~al.(2024)Wu, Wang, Chen, and Carneiro]{wu2024deep}
Renjie Wu, Hu~Wang, Hsiang-Ting Chen, and Gustavo Carneiro.
\newblock Deep multimodal learning with missing modality: A survey.
\newblock \emph{arXiv preprint arXiv:2409.07825}, 2024.

\bibitem[Xu et~al.(2024)Xu, Chen, Du, Zhang, {\L}ukasik, Zhu, and Yu]{xu2024m3a}
Qingzheng Xu, Huiqiang Chen, Heming Du, Hu~Zhang, Szymon {\L}ukasik, Tianqing Zhu, and Xin Yu.
\newblock M3a: A multimodal misinformation dataset for media authenticity analysis.
\newblock \emph{Computer Vision and Image Understanding}, 249:\penalty0 104205, 2024.

\bibitem[Xu and Wang(2017)]{xu2017dynamic}
Shuliang Xu and Junhong Wang.
\newblock Dynamic extreme learning machine for data stream classification.
\newblock \emph{Neurocomputing}, 238:\penalty0 433--449, 2017.

\bibitem[Xue and Marculescu(2023)]{xue2023dynamic}
Zihui Xue and Radu Marculescu.
\newblock Dynamic multimodal fusion.
\newblock In \emph{Proceedings of the IEEE/CVF Conference on Computer Vision and Pattern Recognition}, pages 2575--2584, 2023.

\bibitem[Yang et~al.(2024)Yang, Lu, and Yu]{yang2024adapting}
Xiaoyu Yang, Jie Lu, and En~Yu.
\newblock Adapting multi-modal large language model to concept drift from pre-training onwards.
\newblock \emph{arXiv preprint arXiv:2405.13459}, 2024.

\bibitem[Yu et~al.(2022)Yu, Ma, An, and Li]{yu2022bcmf}
Chuanming Yu, Yinxue Ma, Lu~An, and Gang Li.
\newblock Bcmf: A bidirectional cross-modal fusion model for fake news detection.
\newblock \emph{Information Processing \& Management}, 59\penalty0 (5):\penalty0 103063, 2022.

\bibitem[Zhang and Wei(2017)]{zhang2017review}
Dan Zhang and Bin Wei.
\newblock A review on model reference adaptive control of robotic manipulators.
\newblock \emph{Annual Reviews in Control}, 43:\penalty0 188--198, 2017.

\bibitem[Zhang(2021)]{zhang2021pola}
Wenyu Zhang.
\newblock Pola: Online time series prediction by adaptive learning rates.
\newblock In \emph{ICASSP 2021-2021 IEEE International Conference on Acoustics, Speech and Signal Processing (ICASSP)}, pages 3375--3379. IEEE, 2021.

\bibitem[Zhao et~al.(2024)Zhao, Zhang, Ma, and Cheng]{zhao2024survey}
Tianyi Zhao, Liangliang Zhang, Yao Ma, and Lu~Cheng.
\newblock A survey on safe multi-modal learning systems.
\newblock In \emph{Proceedings of the 30th ACM SIGKDD Conference on Knowledge Discovery and Data Mining}, pages 6655--6665, 2024.

\bibitem[Zou et~al.(2019)Zou, Shen, Jie, Zhang, and Liu]{zou2019sufficient}
Fangyu Zou, Li~Shen, Zequn Jie, Weizhong Zhang, and Wei Liu.
\newblock A sufficient condition for convergences of adam and rmsprop.
\newblock In \emph{Proceedings of the IEEE/CVF Conference on computer vision and pattern recognition}, pages 11127--11135, 2019.

\end{thebibliography}
\bibliographystyle{plainnat}

\newpage
\appendix
\appendixpage

\startcontents[sections]
\printcontents[sections]{l}{1}{\setcounter{tocdepth}{2}}

\newpage
\section{More Related Work}\label{app:relw}
\subsection{Concept Drift Detection and Adaptation}
\begin{figure}[!htbp]
    \centering
    \includegraphics[scale=0.5]{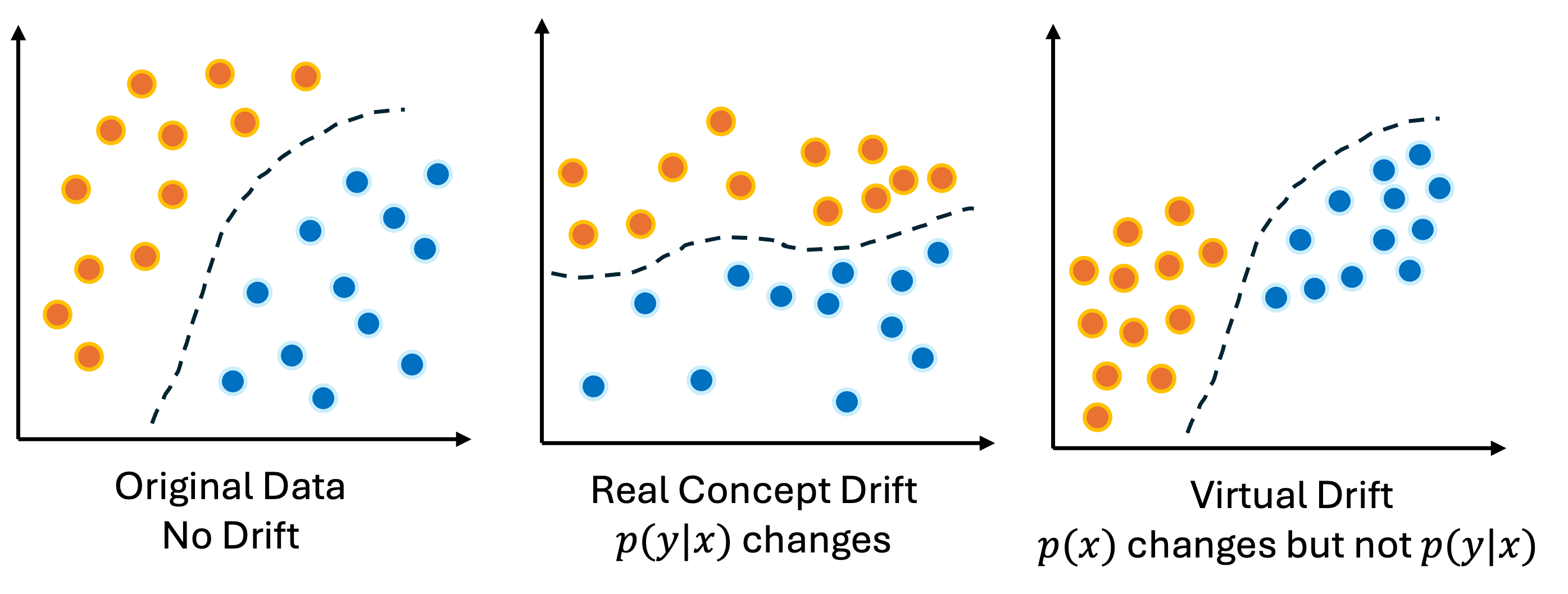}
    \caption{Types of concept drifts: The circles in this figure represent instances, and the different colors represent different classes.}
    \label{fig:concep-drift}
\end{figure}

In machine learning, a "no drift" scenario signifies a stable environment where the joint probability distribution of input features and the target variable, $P(X, Y)$, remains constant over time. Conversely, concept drift occurs when this joint distribution changes, $P_t(X, Y) \neq P_{t'}(X, Y)$ for different time points $t$ and $t'$, and can manifest in sudden, gradual, or incremental patterns. This broader concept can be broken down into two types: real concept drift, which directly impacts predictive modeling by altering the conditional distribution of the target given the input, $P_t(X|Y)$, meaning the fundamental relationship the model learns ($h: X \rightarrow Y$) changes; and virtual drift, where only the input data distribution $P_t(X)$ shifts while the conditional relationship $P_t(Y|X)$ remains unchanged. Figure~\ref{fig:concep-drift} presents the different types of concept drifts. 

\subsection{Adaptive Control Strategies and Learning Rate Adjustment}
Adaptive control provides a framework for real-time adjustment of system parameters to maintain stability and performance. In machine learning, the learning rate is a critical parameter often used with fixed or scheduled values, which can be problematic in concept drift. A low learning rate can hinder the model's ability to learn new concepts, while a high rate may lead to overfitting or instability. Research~\cite{zhang2021pola,zou2019sufficient} indicates that dynamically adapting the learning rate can enhance a model's ability to respond to distribution shifts. For instance, RMSProp and Adam adjust step sizes based on gradient statistics but are not specifically designed for concept drift. Some approaches, like COGRA~\cite{miyaguchi2019cogra}, dynamically tune the learning rate during time-series forecasting by increasing it when prediction error variance rises and decreasing it once performance stabilizes. Similarly, \cite{canonaco2021adaptive} adapts learning rates in federated learning to handle drifting data better. These methods demonstrate the benefits of a higher learning rate post-drift for quicker adaptation. Inspired by these principles, our controller increases the learning rate upon detecting drift to adapt to new data swiftly, then gradually reduces it as the error decreases. This mirrors gain scheduling in control, where one variable (error rate) influences another (learning rate) to optimize performance.

In addition to learning rates, the field of adaptive control presents concepts such as model reference adaptive control (MRAC)~\cite{zhang2017review} and Lyapunov-based adaptation laws~\cite{sun2021lyapunov}, which ensure stability for dynamic systems with unknown parameters. In recent research that combines control theory and deep learning, \cite{patil2024lyapunov} developed Lyapunov stable weight update rules for deep neural network controllers. Although their research focuses on a robotic control loop, the key insight is that parameter update equations can be designed to reduce a Lyapunov function gradually. Our paper treat the learning process itself as a system that needs to be controlled. Additionally, there's a relevant concept in which some researchers have utilized PID control for optimizer tuning~\cite{doicin2016pid}; they treat the training error like a process variable and employ PID controllers to adjust hyperparameters or even the magnitudes of gradients. Our method can be viewed as a specialized version of this approach: when the error increases, the rate of change of the error prompts a proportional rise in the learning rate, similar to a P/PI controller aiming to minimize the error.

\subsection{Lyapunov Stability in Learning Systems}
A Lyapunov function decreases over time concerning the system's state, guaranteeing that it will not diverge from its equilibrium position. In control theory, this provides strong assurances about both convergence and robustness. However, applying Lyapunov stability analysis to machine learning systems is relatively new~\cite{janakiraman2013lyapunov}. Traditional training methods often lack formal stability guarantees, and neural network training can face issues such as exploding gradients or oscillations when hyperparameters are not chosen carefully. Recent efforts have begun to integrate stability theory into learning algorithms. For example, \cite{janakiraman2013lyapunov} proposed a Lyapunov-based online learning algorithm for nonlinear systems, ensuring the model's prediction error remains bounded. Researchers have incorporated Lyapunov-based constraints in deep reinforcement learning to maintain stability in an agent's value function or dynamics during the learning process~\cite{chu2024lyapunov}. 

In adaptive neural controllers, where a neural network controls a physical system, attempts have been made to enforce Lyapunov stability of the closed-loop system. \cite{dai2021lyapunov} developed a neural network policy alongside a neural Lyapunov function that certifies stability within a certain region. This was a verification problem where a mixed-integer program checks the Lyapunov conditions for the neural policy. Such research confirms that it is feasible to train or constrain neural networks while considering Lyapunov criteria. However, these studies mainly focus on controlling external processes, whereas we emphasize the stability of the learning algorithm and its fusion mechanism.

The key innovation in our work is the application of Lyapunov stability analysis to adapt to concept drift in multimodal learning. We construct a Lyapunov function based on the dynamics of the model's error. Intuitively, the error signal parallels the state of a dynamical system. Our adaptive update rules for the learning rate and fusion weight are designed to ensure this error diminishes over time, even in the face of external disturbances caused by drift. This approach is reminiscent of MRAC, where adaptive updates drive the difference between a system and a reference model to zero~\cite{patil2024lyapunov}. In this context, the "reference" is the new concept (or zero error), and we adapt parameters to track it. To the best of our knowledge, no prior work in multimodal learning has incorporated a Lyapunov stability perspective to manage streaming non-stationary data. 

\newpage
\section{Proposed Algorithm}\label{app:algo}
\subsection{LS-OGD Main Adaptation Loop}
The algorithm of the main adaptation loop for multimodal learning is
\begin{algorithm}[!htbp]
\caption{LS-OGD Main Adaptation Loop for Multimodal Learning}
\label{alg:ls_ogd_main_loop}
\begin{algorithmic}[1]
\State \textbf{Input:}
\State \quad Data stream for Phase 2: $\mathcal{D}_{drift} = \{(X_{text}^{(i)}, X_{img}^{(i)}, Y_{true}^{(i)})\}_{i=1}^{N_{P2}}$
\State \quad Model $M_{\theta}$ (Text Encoder $M_T$, Image Encoder $M_I$, Fusion $M_F$) initialized from Phase 1
\State \quad Initial learning rate $\eta_{current}$
\State \quad Initial raw fusion parameter $\hat{\alpha}_{current}$ (for $M_F$)
\State \quad Optimizer $Opt$ (e.g., AdamW) initialized with parameters of $M_{\theta}$
\State \quad Accuracy history deque $H_{acc}$ (max length $L_{hist}$), potentially primed
\State \quad Drift detection parameters: $W_{drift}, R_{comp}, \tau_{drop}$ (passed to Algorithm \ref{alg:ls_ogd_controller_logic})
\State \quad LR adaptation parameters: $k_{lr}, [\eta_{min}, \eta_{max}], \theta_{high\_drift}, \theta_{mod\_drift}$ (passed to Algorithm \ref{alg:ls_ogd_controller_logic})
\State \quad Alpha adaptation parameters: $k_{\alpha}, \text{drift type indicators}$ (passed to Algorithm \ref{alg:ls_ogd_controller_logic})
\State \quad Loss function $\mathcal{L}_{BCE}$ (Binary Cross-Entropy)

\vspace{0.2cm}
\State \textbf{Initialize:} Apply $\eta_{current}$ to $Opt$.

\vspace{0.2cm}
\For{each batch $(X_{text}, X_{img}, Y_{true})$ from $\mathcal{D}_{drift}$}
    \State // Perform forward pass
    \State $p_{text} \gets M_T(X_{text})$ \Comment{Text modality probability}
    \State $p_{img} \gets M_I(X_{img})$ \Comment{Image modality probability}
    \State $\alpha_{display} \gets \sigma(\hat{\alpha}_{current})$ \Comment{Current sigmoid-squashed fusion weight}
    \State $p_{fused} \gets (1 - \alpha_{display}) \cdot p_{text} + \alpha_{display} \cdot p_{img}$ \Comment{Fused probability}

    \vspace{0.1cm}
    \State // Evaluate performance
    \State $L_t \gets \text{mean}(\mathcal{L}_{BCE}(p_{fused}, Y_{true}))$
    \State $Acc_t \gets \text{ComputeAccuracy}(p_{fused}, Y_{true})$
    \State Add $Acc_t$ to $H_{acc}$

    \vspace{0.1cm}
    \State // Obtain controller actions and new parameters from LS-OGD Controller Logic (Algorithm \ref{alg:ls_ogd_controller_logic})
    \State $(S_{drift}, \eta_{new}, \hat{\alpha}_{new}) \gets \text{LSOGDControllerLogic}(H_{acc}, \eta_{current}, \hat{\alpha}_{current}, \text{params})$
    \Comment{params include all detection and adaptation hyperparameters}

    \vspace{0.1cm}
    \State // Apply adaptations and optimize
    \State Update learning rate in $Opt$ to $\eta_{new}$
    \State $Opt.\text{zero\_grad}()$
    \State $L_t.\text{backward}()$
    \State $Opt.\text{step}()$ \Comment{Updates $M_{\theta}$ using $\eta_{new}$}

    \vspace{0.1cm}
    \State // Update state for next iteration
    \State $\eta_{current} \gets \eta_{new}$
    \State $\hat{\alpha}_{current} \gets \hat{\alpha}_{new}$ \Comment{$\hat{\alpha}$ in $M_F$ is now updated to $\hat{\alpha}_{new}$}

    \vspace{0.1cm}
    \State LogMetrics($L_t, Acc_t, \eta_{current}, \sigma(\hat{\alpha}_{current}), S_{drift}$, etc.)
\EndFor
\end{algorithmic}
\end{algorithm}

For each batch, it performs a forward pass using the current fusion weight, evaluates performance, calls Algorithm~\ref{alg:ls_ogd_controller_logic} to get controller actions (drift signal, new learning rate, new fusion parameter), applies these adaptations, performs backpropagation and optimization, and updates the state for the next iteration. Algorithm~\ref{alg:ls_ogd_controller_logic} details the controller's decision-making. It takes the accuracy history and current parameters as input. It consists of three main procedures: calculating a drift signal based on the drop in accuracy between recent and past windows, adjusting the learning rate based on the drift signal strength, with clamping to predefined min/max values, and modifying the raw fusion parameter $\hat{\alpha}$ based on the drift signal and indicators of which modality might be drifting. 

\subsection{LS-OGD Controller Logic}
The proposed LS-OGD controller is
\begin{algorithm}[!htbp]
\caption{LS-OGD Controller Logic}
\label{alg:ls_ogd_controller_logic}
\begin{algorithmic}[1]
\State \textbf{Input for Controller Logic function \textsc{LSOGDControllerLogic}:}
\State \quad Accuracy history deque $H_{acc}$
\State \quad Current learning rate $\eta_{in}$
\State \quad Current raw fusion parameter $\hat{\alpha}_{in}$
\State \quad Drift detection parameters: $W_{drift}, R_{comp}, \tau_{drop}$
\State \quad LR adaptation parameters: $k_{lr}, [\eta_{min}, \eta_{max}], \theta_{high\_drift}, \theta_{mod\_drift}$
\State \quad Alpha adaptation parameters: $k_{\alpha}, \text{drift type indicators}$

\vspace{0.2cm}
\State // Detect concept drift
\State $S_{drift} \gets \text{EstimateDriftSignal}(H_{acc}, W_{drift}, R_{comp}, \tau_{drop})$

\vspace{0.1cm}
\State // Determine new learning rate
\State $\eta_{out} \gets \text{AdaptLearningRate}(\eta_{in}, S_{drift}, k_{lr}, [\eta_{min}, \eta_{max}], \theta_{high\_drift}, \theta_{mod\_drift})$

\vspace{0.1cm}
\State // Determine new raw fusion parameter
\State $\hat{\alpha}_{out} \gets \text{AdaptFusionParameter}(\hat{\alpha}_{in}, S_{drift}, k_{\alpha}, \text{drift type indicators})$

\vspace{0.2cm}
\State \Return $(S_{drift}, \eta_{out}, \hat{\alpha}_{out})$

\vspace{0.3cm}
\Procedure{EstimateDriftSignal}{$H_{acc}, W_{drift}, R_{comp}, \tau_{drop}$}
    \If{$|H_{acc}| < W_{drift}$} \Return $0$ \EndIf
    \State $Acc_{recent} \gets \text{mean accuracy over last } \lfloor W_{drift} \cdot R_{comp} \rfloor \text{ entries of } H_{acc}$
    \State $Acc_{past} \gets \text{mean accuracy over } W_{drift} - \lfloor W_{drift} \cdot R_{comp} \rfloor \text{ entries of } H_{acc} \text{ before recent window}$
    \If{$Acc_{past} - Acc_{recent} > \tau_{drop}$}
        \State \Return $(Acc_{past} - Acc_{recent}) / Acc_{past}$ \Comment{Normalized drop as drift signal}
    \Else
        \State \Return $0$
    \EndIf
\EndProcedure

\vspace{0.3cm}
\Procedure{AdaptLearningRate}{$\eta_{prev}, S_{drift}, k_{lr}, [\eta_{min}, \eta_{max}], \theta_{high}, \theta_{mod}$}
    \If{$S_{drift} > \theta_{high}$}
        \State $\eta_{cand} \gets \eta_{prev} \cdot k_{lr}$
    \ElsIf{$S_{drift} > \theta_{mod}$}
        \State $\eta_{cand} \gets \eta_{prev} / k_{lr}$
    \Else
        \State $\eta_{cand} \gets \eta_{prev}$
    \EndIf
    \State \Return $\max(\eta_{min}, \min(\eta_{cand}, \eta_{max}))$
\EndProcedure

\vspace{0.3cm}
\Procedure{AdaptFusionParameter}{$\hat{\alpha}_{prev}, S_{drift}, k_{\alpha}, \text{drift indicators}$}
    \State $\Delta\hat{\alpha} \gets 0$
    \If{$S_{drift} > \theta_{mod\_drift}$ \textbf{and} (image modality suspected based on indicators)}
        \State $\Delta\hat{\alpha} \gets -k_{\alpha}$ \Comment{Adjust to reduce reliance on suspected drifting image modality}
    \ElsIf{$S_{drift} > \theta_{mod\_drift}$ \textbf{and} (text modality suspected based on indicators)}
        \State $\Delta\hat{\alpha} \gets +k_{\alpha}$ \Comment{Adjust to reduce reliance on suspected drifting text modality}
    \EndIf
    \State \Return $\hat{\alpha}_{prev} + \Delta\hat{\alpha}$ \Comment{Apply clamping if $\hat{\alpha}$ has bounds}
\EndProcedure

\end{algorithmic}
\end{algorithm}

\newpage
\section{Additional Results}\label{app:results}
\subsection{Example 1. Scalar Error Dynamics}
Suppose at time $t$, the error is $e_t$. An adaptive learning step aims to reduce this error. Let the change due to learning be $-\eta_t \lambda e_t$ (where $\lambda>0$ reflects learning efficacy and $\eta_t$ is the learning rate). Let concept drift introduce an additive error $\delta_t$. So, $e_{t+1}=e_t-\eta_t \lambda e_t+\delta_t=(1-$ $\left.\eta_t \lambda\right) e_t+\delta_t$.

Consider $V(t)=\frac{1}{2} e_t^2$. Then,
\begin{align}
\Delta V(t) &= \frac{1}{2}\left(e_{t+1}^2-e_t^2\right)=\frac{1}{2}\left(\left(\left(1-\eta_t \lambda\right) e_t+\delta_t\right)^2-e_t^2\right) \\
&=\frac{1}{2}\left(\left(1-\eta_t \lambda\right)^2 e_t^2+2\left(1-\eta_t \lambda\right) e_t \delta_t+\delta_t^2-e_t^2\right) \\
&=\frac{1}{2}\left(-2 \eta_t \lambda+\eta_t^2 \lambda^2\right) e_t^2+\left(1-\eta_t \lambda\right) e_t \delta_t+\frac{1}{2} \delta_t^2
\end{align}
If $\eta_t \lambda$ is small (e.g., $<1$ ), and we approximate $1-\eta_t \lambda \approx 1$:
\begin{align}
    \Delta V(t) \approx-\eta_t \lambda e_t^2+e_t \delta_t+\frac{1}{2} \delta_t^2.
\end{align}

Our controller (Algorithm~\ref{alg:ls_ogd_controller_logic}) adapts $\eta_t$. Suppose if $\left|e_t\right|$ is persistently large, $\eta_t$ is increased. If $\left|\delta_t\right| \leq \delta_{\max }$ (Axiom A1). For $\Delta V(t)$ to be negative when $\left|e_t\right|$ is large, we need $\eta_t \lambda e_t^2$ to dominate $\left|e_t \delta_t+\frac{1}{2} \delta_t^2\right|$. This can be ensured if $\left|e_t\right|>\frac{\delta_{\max }}{\eta_k \lambda}+\frac{\delta_{\max }^2}{2 \eta_k \lambda\left|e_t\right|}$.

This implies that $e_t$ will be driven towards a bound related to $\delta_{\max } /\left(\eta_t \lambda\right)$. If $\delta_t \rightarrow 0$, then $\Delta V(t) \approx-\eta_t \lambda e_t^2<0$ for $e_t \neq 0$, leading to $e_t \rightarrow 0$. This simplified example illustrates the core mechanism behind UUB and asymptotic convergence. The adaptive controller's role in adjusting $\eta_t$ is crucial for maintaining the dominance of the negative term when $e_t$ is large. 

\subsection{Corollary 2. System Behavior with Partial Modality Degradation}
If, under the conditions of Theorem 2, modality 1 does not become entirely uninformative but rather its signal-to-noise ratio significantly degrades (making $e_t^{(1), \text{est}}$ consistently higher than $e_t^{(2), \text { est }}$ but not necessarily reflective of pure noise), the fusion weight $\alpha_t$ will still decrease, reducing the influence of the degraded modality 1. The system's error $e_t$ will then converge to a UUB state primarily determined by the more reliable modality 2, potentially augmented by any residual (but down-weighted) useful information from modality 1, or slightly inflated by its downweighted noise.

Theorem 2 discusses the case of a modality becoming essentially uninformative. This corollary extends the intuition to a ``soft" failure. It highlights that adaptive fusion is not just a switch; it can find an intermediate balance if a modality is partially compromised. It still attempts to optimize the fusion based on relative estimated reliabilities. The system reduces reliance rather than completely discarding a modality unless its error contribution becomes overwhelmingly negative. 

\subsection{Remark 2. Practical Estimation of Modality-Specific Errors}
The efficacy of Theorem 2 and its supporting Lemma 2 hinges on the ability to obtain reliable estimates of modality-specific errors $\left(e_t^{(m), \text{est}}\right)$. As outlined in Section~\ref{sec:prel} and Algorithm~\ref{alg:ls_ogd_controller_logic}, this often involves techniques like evaluating hypothetical unimodal predictions. The accuracy of these estimates can be affected by:
\begin{enumerate}[(i)]
    \item Shared representations: If early parts of the model are shared before modality-specific processing, disentangling error contributions can be indirect. 
    \item Complexity of drift: Observing the relative degradation might be challenging if drift affects both modalities correlatedly.
    \item Noise in labels or inputs: This can affect the stability of individual error estimates. 
\end{enumerate}

\subsection{Remark 3. Continuous-Time Analysis vs. Discrete-Time Implementation}
For conceptual clarity and leveraging established mathematical tools, our Lyapunov analysis often refers to continuous-time derivatives or small discrete differences $\Delta V(t)$. The proposed LS-OGD system (Algorithms~\ref{alg:ls_ogd_main_loop} and \ref{alg:ls_ogd_controller_logic}) operates in discrete time steps (batches). The theoretical stability results (UUB) derived from a continuous-time perspective generally translate to discrete-time systems provided that the discrete time steps (effectively related to how frequently parameters $\eta_t, \alpha_t$, and model weights $\theta$ are updated) are sufficiently small. If updates are too large or infrequent relative to the system's or drift's dynamics, discrete-time effects could lead to behaviors not captured by the continuous approximation, including instability even if the continuous analog is stable. This highlights the importance of the learning rate and adaptation gain selection in the practical discrete-time implementation.

\subsection{Example 2. Fusion Weight Dynamics}
Assume $\alpha_t$ is the weight for modality 1 (text), so the fused output is $z_t=\alpha_t z_t^{(1)}+\left(1-\alpha_t\right) z_t^{(2)}$. Suppose at $t_0$, modality 1 starts providing misleading information, leading to a consistently higher estimated error $e_t^{(1), \text { est }}$ compared to modality 2's error $e_t^{(2), \text { est }}$. Let $e_t^{(1), \text { est }}=0.8$ and $e_t^{(2), \text { est }}=0.2$ for $t>t_0$. The adaptation logic for $\alpha_t$ is designed to shift weight away from the modality with higher error. An effective update for $\alpha_t$ might be $\alpha_{t+1}=\operatorname{clip}_{[0,1]}\left(\alpha_t+k_\alpha \cdot\right.$ (signal favoring modality 2 over 1)).

If the signal is proportional to $\left(e_t^{(2), \text { est }}-e_t^{(1), \text { est }}\right)$, then $\Delta \alpha_t=k_\alpha(0.2-0.8)=-0.6 k_\alpha$. If $k_\alpha=0.05$, then $\Delta \alpha_t=-0.03$.
Starting with $\alpha_{t_0}=0.5$ (equal weighting):

\begin{align}
\alpha_{t_0+1}&=0.5-0.03=0.47 \\
\alpha_{t_0+2}&=0.47-0.03=0.44
\end{align}

This trend will continue, reducing $\alpha_t$ until it reaches 0 (due to clipping). At this point, the system relies entirely on modality $2\left(z_t \approx z_t^{(2)}\right)$, having effectively isolated the "faulty" modality 1. This illustrates how Lemma 2's assertion of $\alpha_t$ convergence leads to the outcome in Theorem 2. The implementation in Algorithm~\ref{alg:ls_ogd_controller_logic} uses $\hat{\alpha}$ and a sigmoid, but the principle of shifting away from the higher-error modality is the same.

\newpage
\section{Proofs}\label{theoremproof1}
\subsection{Proof for Lemma 1}
By the joint learning-control update (Eq. (5) in the main paper), the prediction error evolves as 
\begin{align}
    e_{t+1} = e_t - \eta_t g(e_t, \alpha_t) + \delta_t + r_t + \rho_t,
\end{align}
where $g(e_t, \alpha_t)$ is the effective gradient-like term from the controller and learner, $\delta_t \in \mathbb{R}^n$ is the bounded drift disturbance (Axiom 1: $||\delta_t|| \leq \delta_{max}$), $r_t$ collects higher-order Taylor-remainder terms in $\eta_t$ (hence, $|r_t| = \mathcal{O}(\eta_t^2)$), and $\rho_t$ collects the effect of the fusion-weight update (hence, $|\rho_t| = \mathcal{O}(||\Delta \alpha_t||)$). 

We defined the Lyapunov function in Eq. (6) and get
\begin{align}
    \Delta V(t) = \frac{1}{2}(e^2_{t+1} - e^2_t) = \frac{1}{2}\Big[(e_t - \eta_t g + \delta_t + \rho_t)^2 - e_t^2\Big].
\end{align}

Expand $e_{t+1}^2$:
\begin{align}
    e_{t+1}^2 &= e_t^2 - 2\eta_t e_t g(e_t,\alpha_t) + 2 e_t \delta_t + 2 e_t (r_t + \rho_t) \\
    &+ (\eta_t g(e_t, \alpha_t))^2 + ||\delta_t||^2 + (r_t + \rho_t)^2 - 2\eta_t g(e_t, \alpha_t)^T(\delta_t + r_t + \rho_t).
\end{align}

Subtracting $e_t^2$ and dividing by 2 gives
\begin{align}
    \Delta V(t) &= -\eta_t e_t g\left(e_t, \alpha_t\right)+e_t \delta_t+e_t\left(r_t+\rho_t\right)+\frac{1}{2} \eta_t^2\left\|g\left(e_t, \alpha_t\right)\right\|^2 \\ 
    &+\frac{1}{2}\left\|\delta_t\right\|^2+\frac{1}{2}\left(r_t+\rho_t\right)^2 -\eta_t g\left(e_t, \alpha_t\right)^{\top}\left(\delta_t+r_t+\rho_t\right).
\end{align}

By Axiom 2, there is $l_1 > 0$ so that
\begin{align}
    e_t g(e_t, \alpha_t) \geq l_1 e_t^2,
\end{align}
and $||g(e_t, \alpha_t)|| \leq l_2 |e_t|$ for some $l_2 > 0$. The remainder $r_t = \mathcal{O}(\eta_t^2)$ and fusion-weight effect $\rho_t = \mathcal{O}(||\Delta \alpha_t||)$, so their contributions can be gathered into 
\begin{align}
    \mathcal{G}(e_t,\alpha_t) = ||g(e_t, \alpha_t)||^2 + \frac{(r_t + \rho_t)^2}{\eta_t^2} + \frac{g(e_t, \alpha_t)^T(r_t+\rho_t)}{\eta_t},
\end{align}
which is bounded by a polynomial in $|e_t|$, $||\Delta \alpha_t||$, and $\eta_t$. By choosing $k_{eta}$, $k_{\alpha}$ small enough, all higher-order terms and cross-terms involving $\delta_t,\, r_t,\, \rho_t$ can be upper-bounded by constants times $\eta_t^2\mathcal{G}(e_t, \alpha_t)$ and $||\Delta \alpha_t||$.

Group the dominant negative term and the rest into the form
\begin{align}
    \Delta V(t) \leq-\underbrace{\ell_1}_{\gamma_1} \eta_t e_t^2+\underbrace{1}_{\gamma_2} \eta_t\left|e_t\right|\left\|\delta_t\right\|+\underbrace{\frac{1}{2}\left(1+\ell_2^2\right)}_{\gamma_3} \eta_t^2\left(\mathcal{G}\left(e_t, \alpha_t\right)+\left\|\delta_t\right\|^2\right)+ \mathcal{O}\left(\left\|\Delta \alpha_t\right\|\right),
\end{align}

Hence, the stated bound holds with $\gamma_1=l_1$, $\gamma_2=1$, $\gamma_3=\frac{1}{2}(1+l_2^2)$, and with the $\mathcal{O}(||\Delta \alpha_t||)$ term capturing the fusion-weight adaptation effect. This completes the proof of Lemma 1.

\subsection{Proof for Lemma 2}
By design (Eq. (8) in the paper), the fusion weight is adjusted each step according to the signed error gap:
\begin{align}
    \alpha_{t+1}=\text{clip}_{[0,1]}\left(\alpha_t-k_\alpha\left(\hat{e}_t^{(1)}-\hat{e}_t^{(2)}\right)\right)=\text{clip}_{[0,1]}\left(\alpha_t-k_\alpha \Delta e_t\right),
\end{align}
where $\text{clip}_{[0,1]}(\cdot)$ denotes clipping into [0,1]. Since by hypothesis $\Delta e_t > 0$ and $k_{\alpha}\Delta e_t < 1$, we have 
\begin{align}
    \alpha_t - k_{\alpha} \Delta e_t < \alpha_t,
\end{align}
and furthermore $\alpha_t - k_{\alpha} \Delta e_t \geq 0$ as long as $\alpha_t \geq k_{\alpha} \Delta e_t$. Because $k_{\alpha}$ is chosen small enough, for any interior $\alpha_t \in (0,1]$ we remain in [0,1] after subtracting $k_{\alpha}\Delta e_t$. Thus, the projection does nothing and 
\begin{align}
    \alpha_{t+1}=\alpha_t-k_\alpha \Delta e_t<\alpha_t.
\end{align}

Hence, whenever $\alpha_t>0$, the update yields $\alpha_{t+1}<\alpha_t$. By induction, the sequence $\left\{\alpha_t\right\}$ is strictly decreasing until it first hits a point $\alpha_T \leq k_\alpha \Delta e_T$. At that step, the unclipped value $\alpha_T-k_{\alpha} \Delta e_T$ would be negative, so the projection forces $\alpha_{T+1}=0$. Thereafter, $\alpha_t \equiv 0$ for all $t \geq T+1$.

Under a persistent positive error gap $\Delta e_t > 0$, the fusion weight moves monotonically downward at rate $k_{\alpha}\Delta e_t$ and converges in finitely many steps to the boundary $\alpha = 0$, fully down-weighting the degraded modality. This completes the proof of Lemma 2.

\subsection{Proof for Proposition 1}
In the interior $\alpha_t \in (0,1)$ and under the persistent gap assumption $\Delta e_t >0$ with $k_{\alpha} \Delta e_t < 1$, the projection $\text{clip}_{[0,1]}$ is inactive. Hence, the update rule 
\begin{align}
    \alpha_{t+1} = \alpha_t - k_{\alpha}\Delta e_t
\end{align}
holds exactly. Taking expectations conditioned on the history up to time $t$ gives
\begin{align}
    \mathbb{E}\left[\alpha_{t+1} \mid \mathcal{F}_t\right]=\alpha_t-k_\alpha \mathbb{E}\left[\Delta e_t \mid \mathcal{F}_t\right].
\end{align}

Since $\alpha_t$ is $\mathcal{F}_t$-measurable,
\begin{align}
    \mathbb{E}\left[\alpha_{t+1}-\alpha_t \mid \mathcal{F}_t\right]=-k_\alpha \mathbb{E}\left[\Delta e_t \mid \mathcal{F}_t\right].
\end{align}

Removing the conditioning yields 
\begin{align}
    \mathbb{E}\left[\alpha_{t+1}-\alpha_t\right]=-k_\alpha \mathbb{E}\left[\Delta e_t\right].
\end{align}

Thus, on average, the fusion weight decreases by a constant amount $k_\alpha \mathbb{E}\left[\Delta e_t\right]$ each step. Denote this mean decrement by
\begin{align}
    d=k_\alpha \mathbb{E}\left[\Delta e_t\right]>0.
\end{align}

Then $\mathbb{E}\left[\alpha_{t+1}\right]=\mathbb{E}\left[\alpha_t\right]-d$, so by unrolling the expectation,
\begin{align}
    \mathbb{E}\left[\alpha_t\right]=\alpha_0-t d.
\end{align}

To reach a small threshold $\varepsilon \in [0, \alpha_t)$, solve
\begin{align}
    \alpha_0 - Td &\leq \varepsilon \\
    T &\geq \frac{\alpha_0-\varepsilon}{d}=\frac{\alpha_0-\varepsilon}{k_\alpha \mathbb{E}\left[\Delta e_t\right]}.
\end{align}

Hence, the expected number of steps to drive the fusion weight down to $\varepsilon$ scales inversely with the product $k_{\alpha}\mathbb{E}[\Delta e_t]$. Because the fusion-weight update is a simple decrement by $k_{\alpha}\Delta e_t$ in expectation, the convergence toward the boundary is linear in expectation, with rate constant $k_{\alpha}\mathbb{E}[\Delta e_t]$. This completes the proof of Proposition 1.

\subsection{Proof for Theorem 1}
We defined the Lyapunov function in Eq. (6). By Lemma 1, there exist constants $\gamma_1, \, \gamma_2, \, \gamma_3 > 0$ and functions $\mathcal{G}(e_t, \alpha_t)$, $\Delta \alpha_t$ such that
\begin{align}
    \Delta V(t) = V(t+1) - V(t) \leq-\gamma_1 \eta_t e_t^2+\gamma_2 \eta_t\left|e_t\right|\left\|\delta_t\right\|+\gamma_3 \eta_t^2\left(\mathcal{G}\left(e_t, \alpha_t\right)+\left\|\delta_t\right\|^2\right)+\mathcal{O}\left(\left\|\Delta \alpha_t\right\|\right),
\end{align}

By Axiom 1, 
\begin{align}
    ||\delta_t|| \leq \delta_{max}.
\end{align}

Also, by design, the controller ensures
\begin{align}
    0<\eta_t \leq \eta_{\max }, \quad\left\|\Delta \alpha_t\right\| \leq \Delta \alpha_{\max},
\end{align}
with both $\eta_{max}$, $\Delta \alpha_{max}$ arbitrarily small when $k_{\eta}$, $k_{\alpha}$ are small. Hence, each of the positive terms on the right-hand side of $\Delta V(t) = V(t+1) - V(t)$ can be bounded as
\begin{align}
    \gamma_2 \eta_t\left|e_t\right|\left\|\delta_t\right\| &\leq \gamma_2 \eta_{\max } \delta_{\max }\left|e_t\right|, \\
    \gamma_3 \eta_t^2\left\|\delta_t\right\|^2 &\leq \gamma_3 \eta_{\max }^2 \delta_{\max }^2 \\
    \mathcal{O}\left(\left\|\Delta \alpha_t\right\|\right) &\leq c_\alpha \Delta \alpha_{\max},
\end{align}
and similarly $\gamma_3 \eta_t^2 \mathcal{G}(e_t, \alpha_t)$ grows at most quadratically in $\eta_{max}$ and $|e_t|$. Thus, there exists a constant $c_1 > 0$ such that
\begin{align}
    \Delta V(t) \leq-\gamma_1 \eta_t e_t^2+c_1\left(\eta_{\max }\left|e_t\right|+\eta_{\max }^2+\Delta \alpha_{\max}\right).
\end{align}

Choose $\varepsilon_B > 0$ and sufficiently small $\eta_{max}$, $\Delta \alpha_{max}$ so that
\begin{align}
    \gamma_1 \eta_{\max } \varepsilon_B^2>c_1\left(\eta_{\max } \varepsilon_B+\eta_{\max }^2+\Delta \alpha_{\max}\right).
\end{align}

Then, whenever $|e_t| > \varepsilon_B$, the negative quadratic term dominates and 
\begin{align}
    \Delta V(t) < 0,
\end{align}
so $V(t)$ and thus $|e_t|$ decreases until $|e_t| \leq \varepsilon_B$, after which it can never exceed $\varepsilon_B$. This is the UUB property. 

Furthermore, if additionally $\delta_t \rightarrow 0$, then for any $\epsilon > 0$, there exists $T$ such that for all $t \geq T$, $||\delta_t|| \leq \epsilon$. Likewise, $\mathcal{G}(e_t, \alpha_t)$ and $||\Delta \alpha_t||$ can be made arbitrarily small by design of the adaptation gains. Hence, for $t$ large enough, the bounded Eq. (36) reduces to 
\begin{align}
    \Delta V(t) \leq-\frac{\gamma_1}{2} \eta_t e_t^2<0, \quad \text{whenever} \; e_t \neq 0.
\end{align}

Thus, $V(t)$ decreases strictly unless $e_t = 0$, forcing $e_t \rightarrow 0$ as $t \rightarrow \infty$. The above establishes the UUB of the error and asymptotic convergence to zero when the drift vanishes, completing the proof of Theorem 1.

\subsection{Proof for Theorem 2}
Theorem 2 is supposing that one modality (say modality 1) experiences a persistent, large drift so that for all $t \geq t_0$
\begin{align}
    \Delta e_t=\hat{e}_t^{(1)}-\hat{e}_t^{(2)}>\delta_{\text{gap}}>0,
\end{align}
where $\delta_\text{{gap}}$ is a fixed positive constant. Under the fusion-weight update of Lemma 2 with step-size $k_{\alpha}$ chosen so that $k_{\alpha}\delta_{\text{gap}} < 1$, the following hold:
\begin{enumerate}
    \item $\alpha_t \rightarrow 0$ in at most $T \leq\left\lceil\frac{\alpha_{t_0}}{k_\alpha \delta_{\text {gap }}}\right\rceil$ steps (finite-time convergence to the boundary). 
    \item Thereafter, $t \geq t_0 + T$, the closed-loop error $e_t$ evolves as a single-modality system using only modality 2 and thus satisfies the UUB and asymptotic convergence properties of Theorem 1 with ``drift" replaced by the residual error of modality 2 alone. 
\end{enumerate}

Since $\Delta e_t \geq \delta_{gap} > 0$ for all $t \geq t_0$, Lemma 2 guarantees 
\begin{align}
    \alpha_{t+1} = \alpha_t - k_{\alpha} \Delta e_t \leq \alpha_t - k_{\alpha} \delta_{gap},
\end{align}
until $\alpha$ first reaches the boundary at 0. By induction, after $m$ steps
\begin{align}
    \alpha_{t_0+m} \leq \alpha_{t_0}-m k_\alpha \delta_{\text{gap}}.
\end{align}

Hence, $\alpha_{t_0+m} \leq 0$ as soon as
\begin{align}
    m \geq \frac{\alpha_{t_0}}{k_\alpha \delta_{\text{gap}}}.
\end{align}

Because $m$ must be an integer, the first time $\alpha$ hits zero is
\begin{align}
    T=\min \left\{m: \alpha_{t_0+m} \leq 0\right\} \leq\left\lceil\frac{\alpha_{t_0}}{k_\alpha \delta_{\text{gap}}}\right\rceil.
\end{align}

At that step, the projection in the update forces $\alpha_{t_0+m} = 0$, and thereafter $\alpha_t \equiv 0$.

For $t \geq t_0+T$, with $\alpha_t=0$, the fusion component relying on modality 1 is entirely disabled. The joint update (Eq. (1) in the paper) then reduces to the single-modality-2 learning-control system:
\begin{align}
    e_{t+1}=e_t-\eta_t g^{(2)}\left(e_t\right)+\delta_t^{(2)}+r_t^{(2)}+\rho_t^{(2)},
\end{align}
where each term now refers only to modality 2 (its drift $\delta^{(2)}_t$, its gradient effect $g^{(2)}$, etc.), and the small remainder $\rho^{(2)}_t$ captures any residual fusion-weight logic (now inactive). 

By the same Lyapunov argument used in Theorem 1, invoking Lemma 1 with the single-modality terms in place of the joint ones, we conclude that:

\textbf{UUB:} There is an $\varepsilon_B^{(2)}$ (determined by the bound on $||\delta^{(2)}_t||$) such that $|e_t|$ enters and remains within $|e_t| \leq \varepsilon_B^{(2)}$.

\textbf{Asymptotic convergence:} If $\delta_t^{(2)} \rightarrow 0$, then $e_t \rightarrow 0$ as $t \rightarrow \infty$.

Thus, once the faulty modality is fully down-weighted, the system behaves like a stable, single-modality learner and inherits all the stability and convergence guarantees of Theorem 1, but now concerning modality 2 alone. This completes the proof of Theorem 2.

\newpage
\section{Additional Experiment Details}\label{app:exp}
We conducted a series of experiments to empirically validate the theoretical contributions of LS-OGD and its capacity to address key challenges in multimodal concept drift. Our setup is designed to simulate real-world scenarios where data distributions change over time, impacting model performance. We compare LS-OGD against a static baseline to demonstrate its adaptive capabilities. The experiments are built upon the M3A misinformation dataset, chosen for its relevance to dynamic, real-world information streams.

\subsection{Dataset and Preprocessing}
\begin{figure}[!htbp]
    \centering
    \includegraphics[scale=0.5]{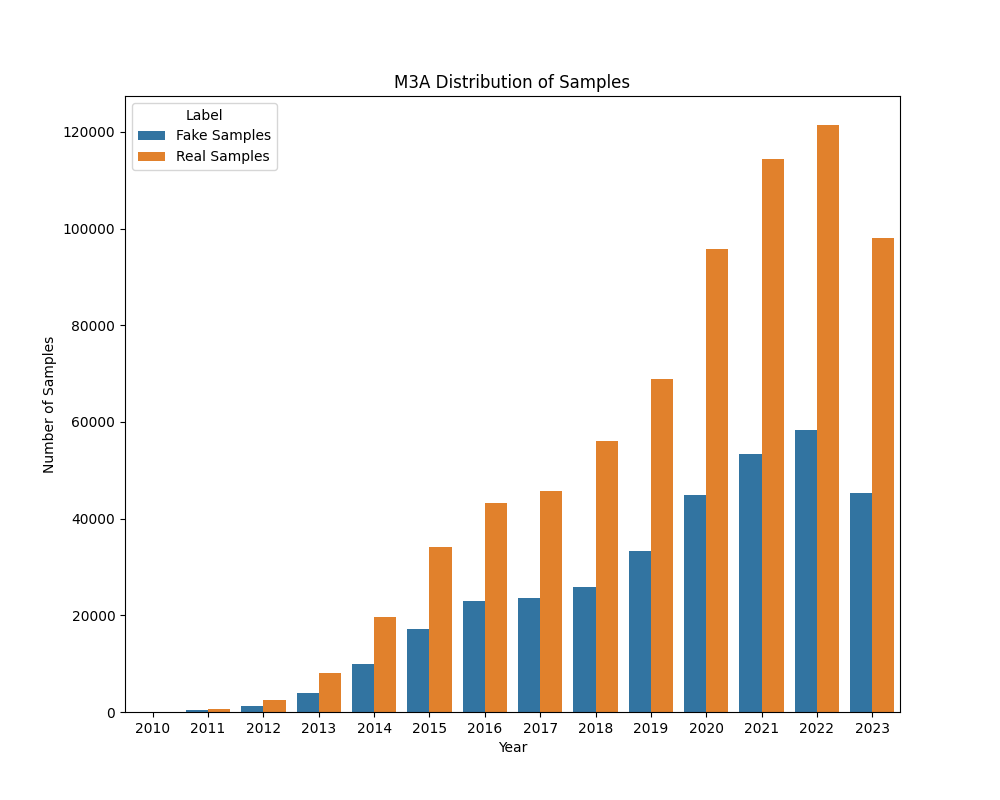}
    \caption{The distribution of samples in the M3A dataset we used in experiments (per year per label)}
    \label{fig:m3a_dis}
\end{figure}

We utilize the M3A dataset, which contains multimodal samples (text and images) relevant to misinformation detection. Table~\ref{tab:dataset-stats} and Figure~\ref{fig:m3a_dis} present the descriptive statistics of the dataset used in this study. 


\begin{table}[!htbp]
  \centering
    \caption{Dataset statistics for text and image modalities (the length of each text sample and the size of each image sample are calculated at the token-level and pixel-level, respectively)}
  \label{tab:dataset-stats}
  \resizebox{\textwidth}{!}{%
    \begin{tabular}{lccc|ccc}
      \toprule
      & \multicolumn{3}{c}{\textbf{Text}} & \multicolumn{3}{c}{\textbf{Images}} \\
      \cmidrule(lr){2-4} \cmidrule(lr){5-7}
      & \# Samples & Max Length & Avg.\ Length & Percentage & Max Size & Avg.\ Size \\
      \midrule
      Real  & 708,425 & 77 & 60.21 & 67.56\% & 2,523,960 & 1,005,724.74 \\
      Fake  & 340,150 & 77 & 38.34 & 32.44\% & 2,523,960 & 1,019,480.79 \\
      \bottomrule
    \end{tabular}%
  }
\end{table}

Following the methodology in this paper, we structure the experiment into two main phases:

\textbf{Phase 1 (Stable Environment):} Data samples are chronologically ordered based on the year column. Samples from years up to and including 2014 are used for initial model training. This phase represents a period of relatively stable data distribution.

\textbf{Phase 2 (Drifting Environment):} Samples from years after 2014 are used to simulate the operational deployment phase where concept drift is introduced.

\subsection{Experimental Setup} 
\subsubsection{Multimodal Model Architecture}
Our multimodal classification model consists of three main components: text encoder, image encoder, and fusion module. We use a pre-trained CLIP-ViT model as the text encoder. The model processes input text to extract semantic features, and a linear head projects these features to a single logit for binary classification. The same ViT model is also employed as the image encoder. Similarly, it processes images to extract visual features, followed by a linear head for logit generation. Regarding the fusion module, the logits from the text and image encoders are combined using a weighted average fusion mechanism, as described in Section~\ref{sec:prel} (Eq. (1) in the paper). This module has a learnable parameter, $\alpha_t$, that balances the contribution of each modality: $z_t = (1-\alpha_t)\cdot z_t^{(\text{text})} + \alpha_t \cdot z_t^{(\text{image})}$.

\subsubsection{Concept Drift Simulation}
In Phase 2, we introduce concept drift into the data stream to test the adaptability of LS-OGD. Based on the configuration, we simulate both image and text modality drifts. For the image modality drift, visual data is degraded by JPEG compression with quality set to 50, addition of Gaussian noise with a standard deviation of 0.05, and application of Gaussian blur with a radius of 3.0. Regarding the text modality drift, textual data undergoes semantic shifts through keyword replacement and appending a fixed phrase to text samples. The configuration allows drift to be applied to all samples or targeted to a specific class.

\subsubsection{Baseline}
We compare LS-OGD against a static baseline. This baseline model is identical to the LS-OGD model architecture but is trained only during Phase 1. In Phase 2, its parameters remain frozen. This baseline represents typical multimodal models that do not adapt to concept drift post-deployment.

\subsubsection{LS-OGD Implementation}
The LS-OGD framework is implemented as an adaptive controller that interacts with the learning process during Phase 2. First, a drift signal is estimated by monitoring the model's accuracy over a sliding window (size of 100 steps). A significant drop in accuracy indicates potential drift. This aligns with the drift detection mechanism described in Section~\ref{sec:prel} of the paper. Second, regarding the learning rate adaptation, if the drift signal exceeds a high threshold (0.7), $\eta_t$ is increased by a factor of 1.5. If it exceeds a moderate threshold (0.2), $\eta_t$ is decreased by the same factor. The learning rate is bounded between 1e-7 and 1e-2. Last, for the fusion weight adaptation, if moderate drift is detected, and the type of simulated drift suggests a particular modality is compromised, the controller adjusts the internal raw logit for $\alpha_t$ by a step of 0.25 to shift reliance towards the more stable modality.

\subsubsection{Training and Evaluation Procedure}
The experiment follows a two-phase procedure:

\textbf{Phase 1 (Initial Training):} The multimodal model (both for what will become the static baseline and the initial state of LS-OGD) is trained on the stable dataset partition for 1000 steps. 

\textbf{Phase 2 (Adaptation and Evaluation under Drift):} For the static baseline evaluation, the model parameters from Phase 1 training are loaded and frozen, and they are used to evaluate on the drifted data from the Phase 2 partition. Regarding LS-OGD adaptation, the model parameters from Phase 1 training are loaded, and the controller is activated. The model continues to train on the drifted data from the Phase 2 partition for 3000 steps, with the controller dynamically adjusting $\eta_t$ and $\alpha_t$. In addition, models are trained using the AdamW optimizer with an initial learning rate of 5e-4 and a weight decay of 0.001. All experiments are seeded for reproducibility.

\subsection{Evaluation Metrics}
To evaluate the performance and characteristics of LS-OGD and the baseline, we track several metrics throughout Phase 2:


\textbf{F1-Score:} The harmonic mean of precision and recall, providing a balanced measure for binary classification. 


\textbf{Loss:} The binary cross-entropy loss. 

\textbf{Learning Rate ($\eta_t$):} The dynamic learning rate applied by LS-OGD.

\textbf{Fusion Alpha ($\alpha_t$):} The dynamic fusion weight used by LS-OGD, indicating reliance on text vs. image modality. 

\textbf{Estimated Drift Signal ($\delta_t$ proxy):} Our drift detection mechanism's output quantifies the perceived change in data distribution based on performance degradation. This is an empirical proxy for $\delta_t$ (Axiom 1 in the paper). 

\textbf{Controller Action:}  The specific adaptation action taken by the LS-OGD controller at each step.

\textbf{Controller Cost:} An assigned cost for each controller action, allowing for analysis of the cumulative cost of adaptation. 

\textbf{Error Signal ($e_t$):} Defined as 1-Accuracy, this is the error signal monitored by the system. 

\textbf{Delta Error Signal ($\Delta e_t$):} The change in the error signal from the previous step, $e_t - e_{t-1}$. This metric serves as an empirical proxy to observe the error dynamics and the behavior of the Lyapunov function candidate $V(t) = \frac{1}{2} \cdot e_t^2$, as discussed in Lemma 1 and Theorem 1 of the paper. 

\begin{figure}[!htb]
    \centering
    \includegraphics[scale=0.38]{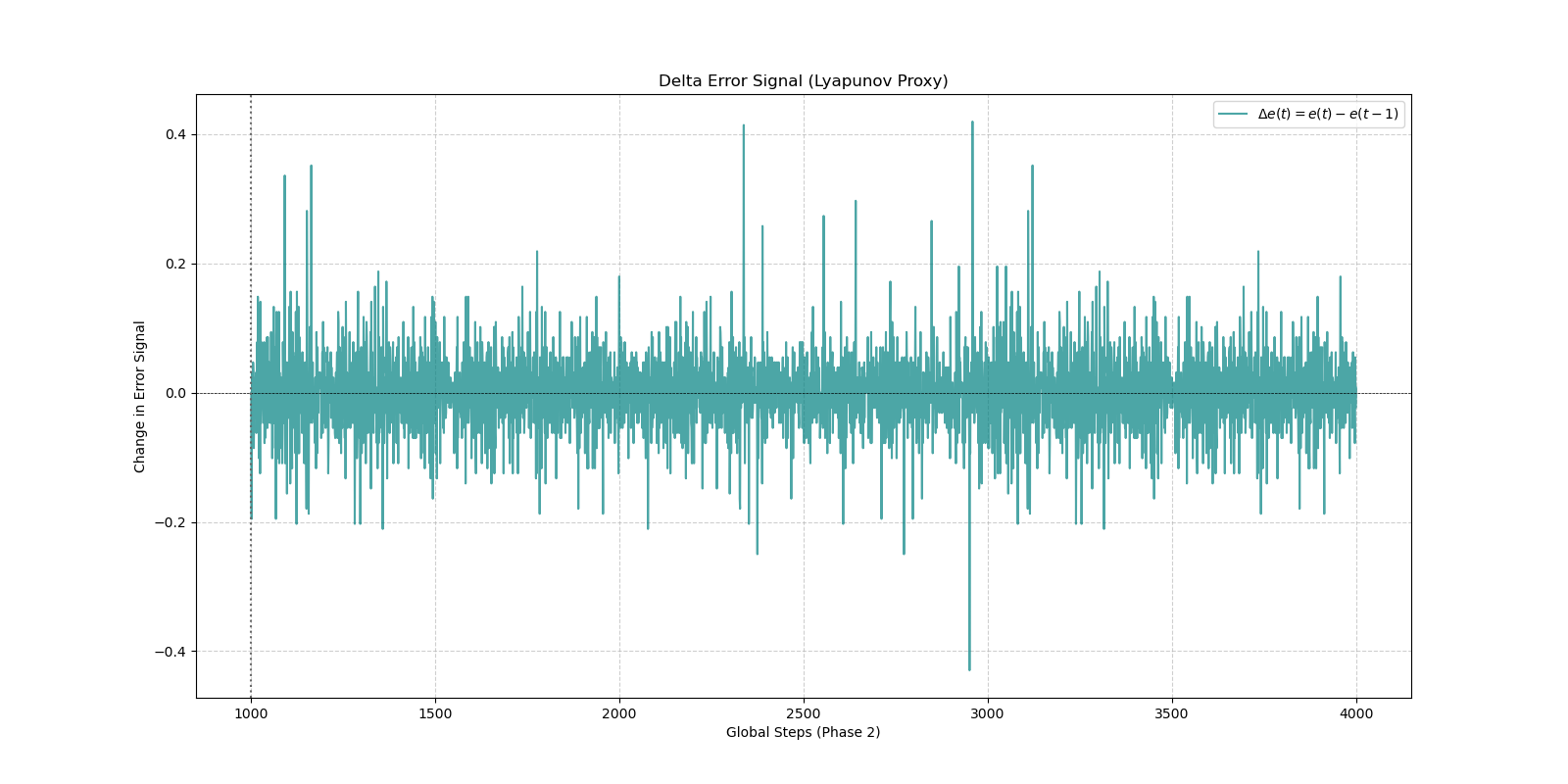}
    \caption{Delta Error Signal for the LS-OGD Model during Phase 2}
    \label{fig:delta_error}
\end{figure}

\subsection{Additional Results}
Our experiments used 4 NVIDIA A100 GPUs with 110 GB of CPU memory per job, which averaged 20 hours of execution.

\begin{figure}[!htb]
    \centering
    \includegraphics[scale=0.38]{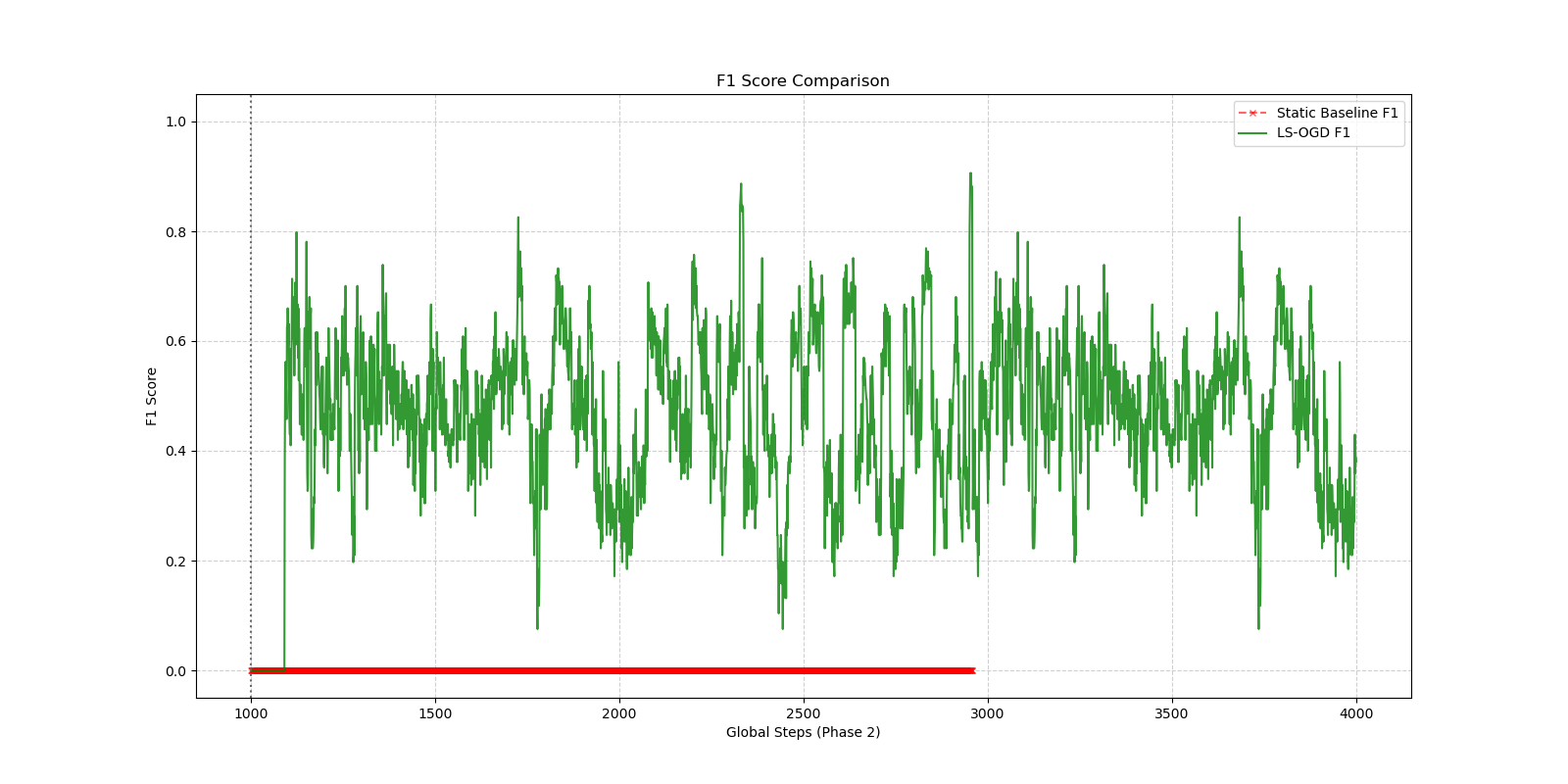}
    \caption{The Comparison between the Static Baseline F1 and LS-OGD F1 over Phase 2}
    \label{fig:f1}
\end{figure}

Figure~\ref{fig:delta_error} is an empirical proxy for observing the behavior related to the Lyapunov stability analysis presented in Lemma 1 and Theorem 1. The Lyapunov function $V(t)=\frac{1}{2}e_t^2$ is stable if its change $\Delta V(t)$ is generally bounded. While this plot shows $\Delta e_t$ rather than $\Delta V(t)$, $\Delta e_t$ frequently becomes negative (error is decreasing), supporting the idea that the adaptive controller effectively drives the error down after perturbations. The bounded nature of these fluctuations suggests that the error is not growing uncontrollably, which aligns with the concept of UUB of the error $e_t$. UUB implies that the error will remain within a certain bound even under persistent drift. Positive spikes in $\Delta e_t$ are expected when concept drift occurs, as the existing model becomes misaligned with the new data distribution, causing an increase in error. The LS-OGD controller's role is to take corrective actions to make subsequent $\Delta e_t$ values negative, thereby reducing the overall error $e_t$. 

\begin{figure}[!htb]
    \centering
    \includegraphics[scale=0.38]{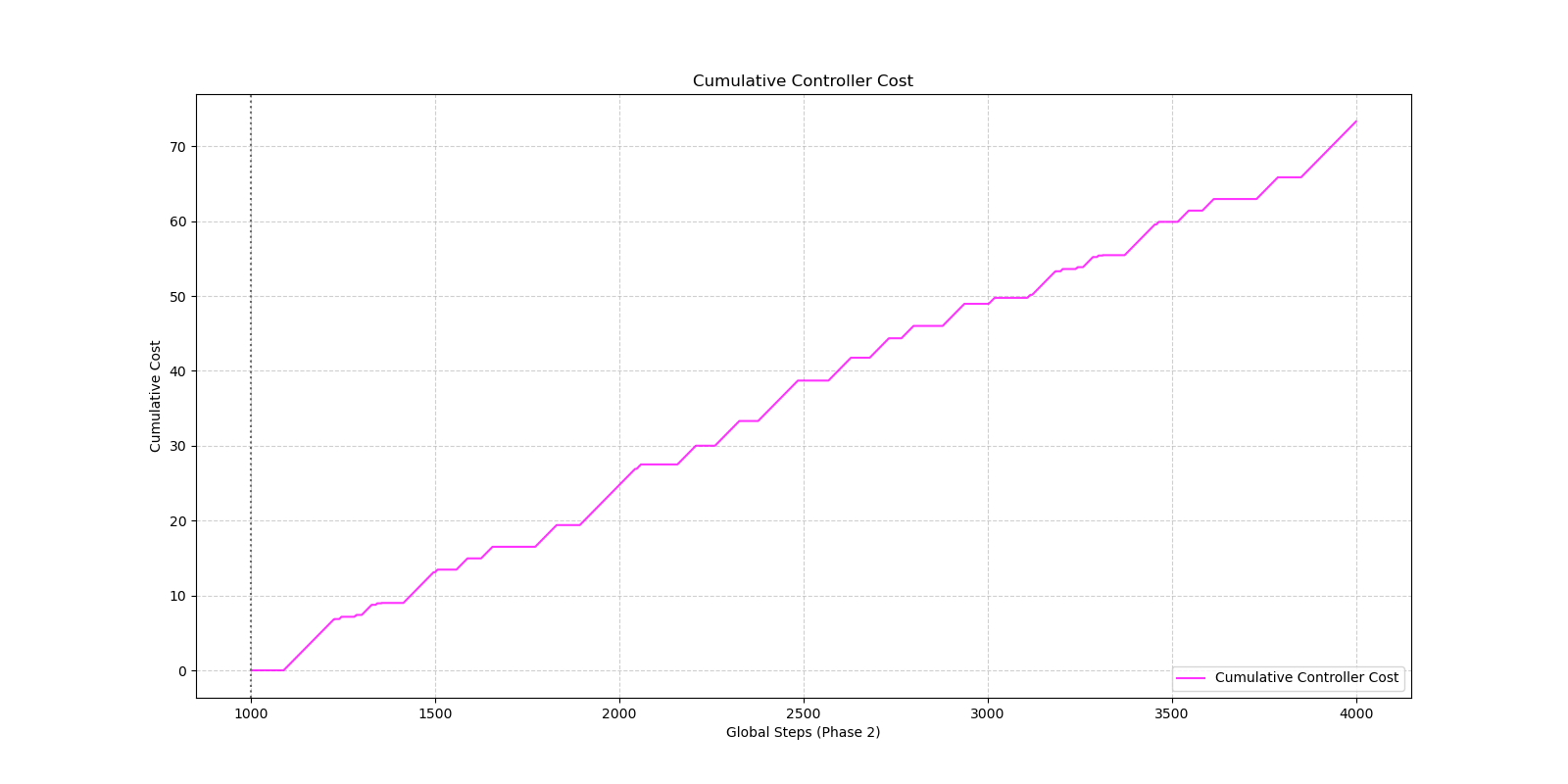}
    \caption{Cumulative Controller Cost for the LS-OGD Model Throughout Phase 2}
    \label{fig:cost}
\end{figure}



Figure~\ref{fig:f1} supports the effectiveness of the LS-OGD framework. The near-zero F1 score of the static baseline highlights the detrimental impact of concept drift on non-adaptive models. LS-OGD maintains a much better F1 score, indicating that it can adapt its decision boundary and modality reliance to changing data. The variability in LS-OGD's F1 score indicates a continuous learning and adaptation process, with the system constantly trying to catch up with the drift. This result aligns with the paper's goal of developing a robust multimodal learning system under concept drift.

Figure~\ref{fig:cost} provides insight into the ``operational overhead" of the LS-OGD adaptive controller. While adaptation is crucial for maintaining performance under drift, it does not come for free. Each adjustment might represent a change in system state that could have other minor costs. The shape of this curve can be correlated with other plots. For instance, periods of rapid increase in controller cost would likely correspond to significant activity in Figure~\ref{fig:adap_sig}, such as changes in learning rate or fusion alpha. Analyzing this cost is important for practical deployments. A controller that adapts effectively but at a high cumulative cost might be less desirable than one that achieves a good balance. The current costs are proxies in our experiments. In a real system, these could represent actual computational overhead, risk associated with change, or resource consumption. This figure helps quantify how ``active" the controller needs to be to manage the specific drift scenario in the experiments.

\newpage
\section{Test-Time Adaptation (TTA) vs. LS-OGD}
Most TTA methods predominantly utilize a fixed, single-modal model for inference, operating without access to labels or source data. These approaches typically focus on minimizing prediction entropy and, in some cases, adjusting normalization layers. The primary objective is to achieve rapid and lightweight adaptation to a stationary or slowly drifting target domain, as exemplified by approaches like Tent~\cite{wang2020tent}. In contrast, the LS-OGD framework addresses the challenges posed by streaming, potentially non-stationary multimodal inputs that exhibit modality-specific drift. It introduces a controller capable of jointly adjusting both the learning rate and the fusion weight in real time. Additionally, LS-OGD offers Lyapunov-style stability guarantees, a feature notably absent in traditional TTA methodologies.

Canonical TTA techniques~\cite{wang2020tent} enhance model confidence through entropy minimization, typically employing only test batches to update batch normalization (BN) affine parameters and/or running statistics. Subsequent variants aimed at stabilizing this objective include CoTTA~\cite{wang2022continual}, which incorporates model and augmentation averaging alongside stochastic weight restoration to mitigate forgetting induced by continual data drift. Additionally, SAR~\cite{niu2023towards} employs filtering of high-gradient and noisy samples while executing sharpness-aware entropy minimization, thereby avoiding performance collapse in dynamic and varied data streams. In contrast, LS-OGD introduces a different adaptive approach by focusing on the adaptation of the global learning rate and a time-varying fusion coefficient, $\alpha(t)$. This method conceptualizes prediction error as a feedback signal within a control loop, utilizing error trends to strategically schedule plasticity and adjust modality weights in response to drift, an adaptation mechanism not addressed by pure entropy objectives.

Furthermore, a significant portion of TTA research focuses on the evaluation of static corruptions or gradually varying sequences. In contrast, when evaluation shifts to continual TTA, methodologies must address challenges such as error accumulation and catastrophic forgetting. The framework of CoTTA~\cite{wang2022continual} has been established to formalize continual TTA, demonstrating that naive updates based solely on entropy tend to degrade performance over time. CoTTA proposes enhancements such as weight and augmentation averaging, along with stochastic restoration, to mitigate these issues. Subsequent studies, including NOTE~\cite{gong2022note}, EcoTTA~\cite{song2023ecotta}, and SoTTA~\cite{gong2023sotta}, have contributed to additional robustness under non-independent and identically distributed (non-IID) streaming data and have also addressed constraints related to memory and noise that deviate from the main interest. LS-OGD presents a framework explicitly designed to tackle bounded yet recurring drift, providing theoretical guarantees concerning error bounds during the presence of drift and ensuring convergence when the drift subsides. This formulation offers a principled alternative to approaches that rely solely on heuristic stabilization methods. 

Moreover, recent advancements in contemporary TTA emphasize empirical stability through various methods, such as sharpness-aware updates for achieving flat minima, entropy ranking/filtering techniques, and protected online self-training combined with change-point detection. However, these approaches still fall short in providing guarantees regarding closed-loop behavior in the presence of drift. The LS-OGD framework introduces a Lyapunov-based analysis that ensures bounded error under drift conditions and demonstrates convergence following instances of drift. This framework allows for the tuning of adaptation laws within stable gain regions, thereby incorporating control-theoretic rigor into the process of online model updates.

\newpage
\section{Limitation and Broader Impact}\label{app:lim}

\textbf{Limitation:}
While LS-OGD offers robust theoretical guarantees and adaptive capabilities, several limitations must be considered. The framework's practical effectiveness, particularly regarding the dynamic adjustments of fusion weights (as supported by Theorem 2), heavily relies on the accuracy of the underlying drift detection mechanism and the reliable estimation of modality-specific error contributions. In highly entangled multimodal drift scenarios, where changes in different modalities are strongly correlated or the individual impacts of errors are difficult to separate, the current controller's error-based attribution logic may face significant challenges.

Furthermore, LS-OGD introduces several control-specific hyperparameters, such as adaptation gains ($k_\eta, k_\alpha$) and various drift detection thresholds (e.g., $W_{\text{drift}}, R_{\text{comp}}, \tau_{\text{drop}}$). Although our theoretical results indicate stability for ``appropriately small" gains, optimal tuning to achieve both rapid adaptation and maximum stability in practice may require careful empirical validation across different datasets and drift characteristics.

Moreover, while Axiom 1 supports our stability proofs, the system's performance and guaranteed stability margins during exceptionally rapid or severe drift events, those that significantly violate this boundedness assumption, need further empirical investigation. Finally, the current adaptation focuses primarily on learning dynamics (via $\eta_t$) and inter-modality reliance (via $\alpha_t$). Future research could explore extending adaptive control principles to the internal architectural components or feature representations within the modality-specific encoders, which may enhance resilience to evolving data characteristics.

Additionally, another limitation of the current framework is its dependence on accurate supervised feedback during the test phase to effectively drive the controller. This approach diverges from optimizing unsupervised criteria, such as prediction entropy, or from stabilizing continual data streams without supervision. Such reliance raises a practical challenge, creating a disconnect between the theoretical guarantees presented in the paper and real-world applications characterized by limited labels. To address this gap, two future research directions are proposed. First, the analysis could be extended to accommodate noisy or weak feedback, thereby allowing the controller to leverage imperfect signals that are prevalent in uncontrolled environments, such as clicks or upvotes. Second, it is advisable to eliminate the dependency on supervisory uncertainty as feedback for the controller. In this context, deep ensembles emerge as a well-validated, label-free proxy for monitoring epistemic uncertainty online, facilitating the detection of shifts and enabling appropriate adaptation strategies.

\textbf{Broader Impact:} 
The development of the LS-OGD framework, which combines control-theoretic stability guarantees with multimodal artificial intelligence (AI), represents a significant step toward creating trustworthy adaptive AI systems that can be more safely deployed in dynamic, real-world environments. Beyond its immediate theoretical contributions, our approach can serve as a blueprint for enhancing continual learning and robust adaptation in domains with inherent non-stationarity and multimodal data. 

In autonomous vehicles (AV) design, ensuring operational reliability under constantly changing conditions is critical. AVs rely on a combination of sensors, and their data characteristics can shift due to weather, lighting, sensor degradation, or novel road scenarios. The LS-OGD's principles of adaptive fusion could enable an AV's perception system to dynamically re-weight sensor inputs, reducing reliance on a modality compromised by fog or glare while prioritizing others. Importantly, Theorem 1 offers a pathway to more predictable and verifiable adaptation behavior. The ability to adapt the learning rate would allow AVs to quickly learn from new, unexpected encounters on the road while maintaining overall system stability.

For national security applications, intelligence analysis frequently involves fusing information from disparate multimodal sources, including satellite imagery, textual reports, signals intelligence, and open-source media. The operational environment is characterized by adversaries actively changing tactics, the emergence of novel threat signatures, and evolving geopolitical landscapes. They all are different forms of concept drift. An LS-OGD-like system could provide robust decision support by adaptively integrating these diverse data streams, automatically down-weighting sources that become outdated, compromised, or intentionally deceptive. The stability guarantees ensure that analytical tools remain reliable even as threat patterns evolve, allowing for more consistent situational awareness and the potential for early detection of novel anomalous activities by adapting to new ``normal" baselines. 

The challenge of misinformation detection also benefits from stable multimodal adaptation. Misinformation campaigns are increasingly sophisticated, blending text, images, and video, and rapidly evolving their narratives, visual styles, and distribution tactics to evade detection. The ability of LS-OGD to adapt its fusion strategy is critical. For example, it could learn to prioritize visual cues if textual components become more anodyne to bypass filters, or vice-versa, as discussed in our problem formulation. By continuously adapting to these shifting tactics with bounded error (Theorem 1), LS-OGD offers a more resilient defense mechanism than static models that quickly become obsolete, thereby contributing to a more informed public discourse.

\newpage
\section{NeurIPS Paper Checklist}

\begin{enumerate}

\item {\bf Claims}
    \item[] Question: Do the main claims made in the abstract and introduction accurately reflect the paper's contributions and scope?
    \item[] Answer: \answerYes{} 
    \item[] Justification: \textcolor{blue}{The abstract and introduction clearly state the claims made, including the contributions made in the paper and important assumptions and limitations.}

\item {\bf Limitations}
    \item[] Question: Does the paper discuss the limitations of the work performed by the authors?
    \item[] Answer: \answerYes{} 
    \item[] Justification: \textcolor{blue}{The limitations of this work have been discussed in our Appendix F.}

\item {\bf Theory assumptions and proofs}
    \item[] Question: For each theoretical result, does the paper provide the full set of assumptions and a complete (and correct) proof?
    \item[] Answer: \answerYes{} 
    \item[] Justification: \textcolor{blue}{All the theorems, formulas, and proofs in this paper have been numbered and cross-referenced. All assumptions are clearly stated and referenced in the statement of any theorems. The proofs are in our Appendix D.}

    \item {\bf Experimental result reproducibility}
    \item[] Question: Does the paper fully disclose all the information needed to reproduce the main experimental results of the paper to the extent that it affects the main claims and/or conclusions of the paper (regardless of whether the code and data are provided or not)?
    \item[] Answer: \answerYes{} 
    \item[] Justification: \textcolor{blue}{Although this paper is submitted to the theory track, we still provide empirical analysis and experiments in both the main paper and Appendix E.}

\item {\bf Open access to data and code}
    \item[] Question: Does the paper provide open access to the data and code, with sufficient instructions to faithfully reproduce the main experimental results, as described in supplemental material?
    \item[] Answer: \answerYes{} 
    \item[] Justification: \textcolor{blue}{Anyone who is interested in reproducing our experiments using M3A data must apply for access from their official GitHub repo: https://github.com/FinalYou/M3A?tab=readme-ov-file. Our code is publicly available online for peer review only at this link: https://github.com/IDontKnowWhoYouAre1022/LS-OGD\_Review\_ONLY}

\item {\bf Experimental setting/details}
    \item[] Question: Does the paper specify all the training and test details (e.g., data splits, hyperparameters, how they were chosen, type of optimizer, etc.) necessary to understand the results?
    \item[] Answer: \answerYes{} 
    \item[] Justification: \textcolor{blue}{All experimental setting and details have been provided in our Appendix E.}

\item {\bf Experiment statistical significance}
    \item[] Question: Does the paper report error bars suitably and correctly defined or other appropriate information about the statistical significance of the experiments?
    \item[] Answer: \answerYes{} 
    \item[] Justification: \textcolor{blue}{The experiment statistics and evaluation metrics have been provided in Appendix E. }

\item {\bf Experiments compute resources}
    \item[] Question: For each experiment, does the paper provide sufficient information on the computer resources (type of compute workers, memory, time of execution) needed to reproduce the experiments?
    \item[] Answer: \answerYes{} 
    \item[] Justification: \textcolor{blue}{The experiments compute resources are reported in Appendix E.}
    
\item {\bf Code of ethics}
    \item[] Question: Does the research conducted in the paper conform, in every respect, with the NeurIPS Code of Ethics \url{https://neurips.cc/public/EthicsGuidelines}?
    \item[] Answer: \answerYes{} 
    \item[] Justification: \textcolor{blue}{The authors have reviewed the NeurIPS Code of Ethics.}

\item {\bf Broader impacts}
    \item[] Question: Does the paper discuss both potential positive societal impacts and negative societal impacts of the work performed?
    \item[] Answer: \answerYes{} 
    \item[] Justification: \textcolor{blue}{We have discussed the broader impacts in the conclusion section and Appendix F.}
    
\item {\bf Safeguards}
    \item[] Question: Does the paper describe safeguards that have been put in place for responsible release of data or models that have a high risk for misuse (e.g., pretrained language models, image generators, or scraped datasets)?
    \item[] Answer: \answerNA{} 
    \item[] Justification: \textcolor{blue}{NA.}

\item {\bf Licenses for existing assets}
    \item[] Question: Are the creators or original owners of assets (e.g., code, data, models), used in the paper, properly credited and are the license and terms of use explicitly mentioned and properly respected?
    \item[] Answer: \answerYes{} 
    \item[] Justification: \textcolor{blue}{We have cited all resources we used in our reference.}

\item {\bf New assets}
    \item[] Question: Are new assets introduced in the paper well documented and is the documentation provided alongside the assets?
    \item[] Answer: \answerNA{} 
    \item[] Justification: \textcolor{blue}{NA.}

\item {\bf Crowdsourcing and research with human subjects}
    \item[] Question: For crowdsourcing experiments and research with human subjects, does the paper include the full text of instructions given to participants and screenshots, if applicable, as well as details about compensation (if any)? 
    \item[] Answer: \answerNA{} 
    \item[] Justification: \textcolor{blue}{NA.}

\item {\bf Institutional review board (IRB) approvals or equivalent for research with human subjects}
    \item[] Question: Does the paper describe potential risks incurred by study participants, whether such risks were disclosed to the subjects, and whether Institutional Review Board (IRB) approvals (or an equivalent approval/review based on the requirements of your country or institution) were obtained?
    \item[] Answer: \answerNA{} 
    \item[] Justification: \textcolor{blue}{NA.}

\item {\bf Declaration of LLM usage}
    \item[] Question: Does the paper describe the usage of LLMs if it is an important, original, or non-standard component of the core methods in this research? Note that if the LLM is used only for writing, editing, or formatting purposes and does not impact the core methodology, scientific rigorousness, or originality of the research, declaration is not required.
    \item[] Answer: \answerNo{} 
    \item[] Justification: \textcolor{blue}{LLMs were only used for editing purposes and do not impact the core methodology, scientific rigor, and originality of this research.}

\end{enumerate}

\end{document}